\documentclass[journal]{IEEEtran}
\usepackage{cite}
\ifCLASSINFOpdf
\usepackage[pdftex]{graphicx}
\else
\fi
\usepackage[numbers,sort&compress]{natbib}
\usepackage{amsmath}
\usepackage{algorithmic}
\usepackage{longtable}
\usepackage{tabu}
\usepackage{multirow}
\usepackage{multicol}
\usepackage{array}
\usepackage{amssymb}
\usepackage{caption}
\ifCLASSOPTIONcompsoc
 \usepackage[caption=false,font=normalsize,labelfont=sf,textfont=sf]{subfig}
\else
\fi
\usepackage{booktabs}
 \usepackage{dblfloatfix}
\usepackage[pdftex,linkcolor=blue,citecolor=blue]{hyperref}
\usepackage{subfigure}  
\usepackage{color}
\usepackage{float}
\begin{document}
\title{Image Feature Information Extraction for Interest Point Detection: A Review}
\author{Junfeng Jing, Tian Gao, Weichuan Zhang, \textsl{Member}, \textsl{IEEE}, Yongsheng Gao, \textsl{Senior Member}, \textsl{IEEE}, Changming Sun

\IEEEcompsocitemizethanks{\IEEEcompsocthanksitem J. Jing and T. Gao are with College
of Electrical and Information, Xi'an Polytechnic University, Xi'an, 710048, China.\protect\\
E-mail: jingjunfeng0718@sina.com; gaotian970228@163.com
\IEEEcompsocthanksitem W. Zhang is with the Institute for Integrated and Intelligent Systems, Griffith University, QLD, Australia, and also with CSIRO Data61, PO Box 76, Epping, NSW 1710, Australia.\protect\\
E-mail: zwc2003@163.com
\IEEEcompsocthanksitem Y. Gao is with the Institute for Integrated and Intelligent Systems, Griffith University, QLD, Australia.\protect\\
E-mail: yongsheng.gao@griffith.edu.au
\IEEEcompsocthanksitem C. Sun is with CSIRO Data61, PO Box 76, Epping, NSW 1710, Australia.\protect\\
E-mail: changming.sun@csiro.au
}

}

\markboth{}%
{Shell \MakeLowercase{\textit{et al.}}: Bare Advanced Demo of IEEEtran.cls for IEEE Computer Society Journals}

\maketitle
\begin{abstract}
Interest point detection is one of the most fundamental and critical problems in computer vision and image processing. In this paper, we carry out a comprehensive review on image feature information (IFI) extraction techniques for interest point detection. To systematically introduce how the existing interest point detection methods extract IFI from an input image, we propose a taxonomy of the IFI extraction techniques for interest point detection. According to this taxonomy, we discuss different types of IFI extraction techniques for interest point detection. Furthermore, we identify the main unresolved issues related to the existing IFI extraction techniques for interest point detection and any interest point detection methods that have not been discussed before. The existing popular datasets and evaluation standards are provided and the performances for eighteen state-of-the-art approaches are evaluated and discussed. Moreover, future research directions on IFI extraction techniques for interest point detection are elaborated.
\end{abstract}
\IEEEpeerreviewmaketitle
\section{Introduction}
\begin{figure*}[!htbp]
\setlength{\abovecaptionskip}{5pt}
\setlength{\belowcaptionskip}{0pt}
\centering
\includegraphics[width=7in]{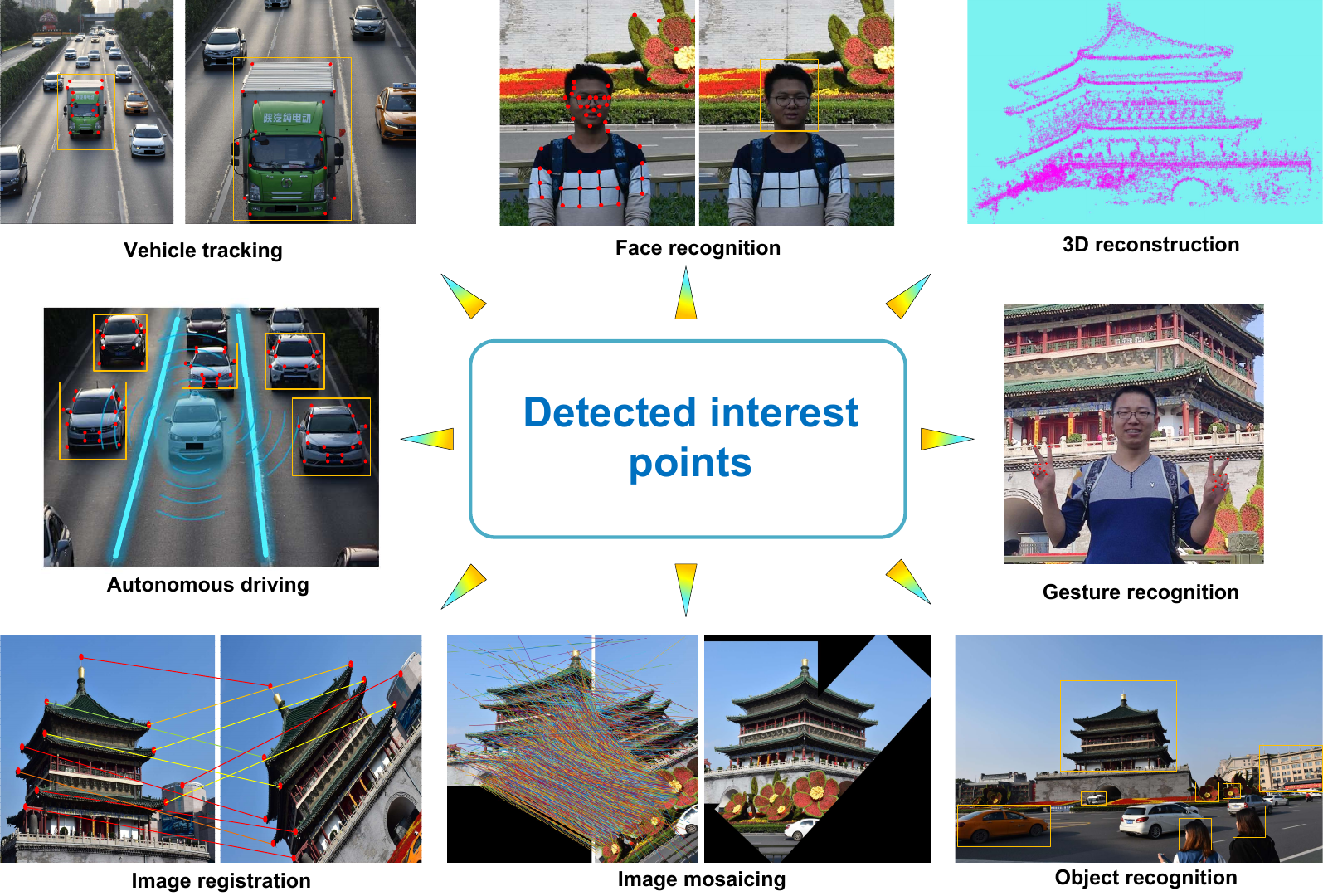}
\caption{Some application examples using detected interest points.}
\label{fig1}
\end{figure*}

\IEEEPARstart{I}{nterest} points have been found very important in human perception of shapes~\cite{attneave1954some} and have been one of the key image features which are widely used as critical cues for various image processing and image understanding tasks such as camera calibration~\cite{zhang1999flexible}, image registration~\cite{ma2018guided}, object recognition~\cite{lowry2015visual}, 3D reconstruction~\cite{fan2019performance}, human-computer interaction~\cite{tang2018Structured}, autonomous driving~\cite{huang2016underwater}, face recognition~\cite{he2018dynamic} and traffic analysis~\cite{datondji2016survey}  as shown in Fig.~\ref{fig1}. Apart from these, the interest point detection techniques are also widely used in visual communication~\cite{pearson1985visual} such as multi-object detection and tracking~\cite{girdhar2018detect}, image-guided robotic-assisted surgery~\cite{mirota2011system}, surveillance~\cite{wang2013intelligent}, and augmented reality ~\cite{amin2015comparative}

Notable interest point detection surveys have been summarized in Table~\ref{stest}. The existing interest point detection surveys include many articles on the problems of specific interest point detections, such as blob detection~\cite{schmid2000evaluation}, local invariant feature detection~\cite{tuytelaars2008local}, interest point detection for visual tracking~\cite{gauglitz2011evaluation}, edge contour based corner detection~\cite{awrangjeb2012performance}, interest point detection for image matching~\cite{miksik2012evaluation}, 3D keypoint detection~\cite{tombari2013performance}, and local feature detection for 3D reconstruction~\cite{fan2019performance}. Our investigation indicates that the existing surveys on interest point detection do not accurately describe the characteristics, advantages, and disadvantages of the interest point detection algorithms, especially those well-known algorithms (e.g., the Harris method~\cite{harris1988combined} and the Harris-Laplace method~\cite{mikolajczyk2004scale}). Moreover, the key issue for image interest point detection is how to directly or indirectly extract and analyze image feature information (IFI) such as first-order, second-order, or high-order intensity variation information, image frequency variation information, image local patterns (e.g., edges or contours), or high level feature representation obtained from a designed machine learning based network architecture. The quality of the IFI extracted from an image directly affects the performance of interest point detection (e.g., the repeatability of interest point detection under image affine transformation). There lacks a comprehensive survey about IFI extraction techniques for interest point detection. Furthermore, there has been a vigorous development of image interest point detection in last decade years. In this paper, we intend to fill this gap, and a taxonomy of the IFI extraction techniques for interest point detection is proposed. According to this classification method, we discuss different types of IFI extraction techniques in depth, and put forward our own views on the different IFI extraction techniques in the literature.

The number of papers on interest point detection is overwhelming. Because of this, compiling any exhaustive review of the approaches is out of the scope of a paper with a reasonable length. As a result, we have limited our focus on 2D image interest point detection methods that were published in top quality journals or conferences and are highly cited. It is worth to note that the existing blob detection techniques include two categories~\cite{Li2015A}: interest point detection~\cite{savinov2017quad} and interest region detection~\cite{Matas2004robust, varytimidis2016alpha}. Considering that our theme is on 2D point like features, this paper only discusses the point feature detection category on blob detection techniques, and it does not discuss the region based blob detection techniques.

The highlights of this article comprise five aspects. First, a taxonomy of the existing IFI extraction techniques for interest point detection is presented. Second, different types of IFI extraction techniques for image interest point detection are analyzed and summarized. Meanwhile, the properties of the existing IFI extraction techniques for interest point detection are analyzed in detail. Third, comprehensive performance evaluations are used for performance testing for eighteen state-of-the-art algorithms. Fourth, the results of the experiments are systemically analyzed and summarized. Fifth, the development trend on IFI extraction for interest point detection is presented.

The remainder of the article is arranged as follows. In Section 2, related background and main challenges on interest point detection are summarized. Section 3 presents a taxonomy of the existing IFI extraction techniques for image interest point detection. In terms of this taxonomy, we discuss different types of IFI extraction techniques in depth and put forward our own views on different IFI extraction techniques in the literature. In Section 4, the existing popular datasets and evaluation standards are summarized and the performances for the eighteen state-of-the-art approaches are discussed. In Section 5, the development trend on IFI extraction for interest point detection is elaborated. Finally, conclusions are presented in Section 6.

\section{Generic Interest Point Detection}
In the following, we firstly provide a brief background on the definition of an interest point. And then the main challenges on interest point detection are presented.

\subsection{Background}
\begin{figure}[!t]

\setlength{\abovecaptionskip}{5pt}
\setlength{\belowcaptionskip}{5pt}
\centering
\includegraphics[width=3.5in]{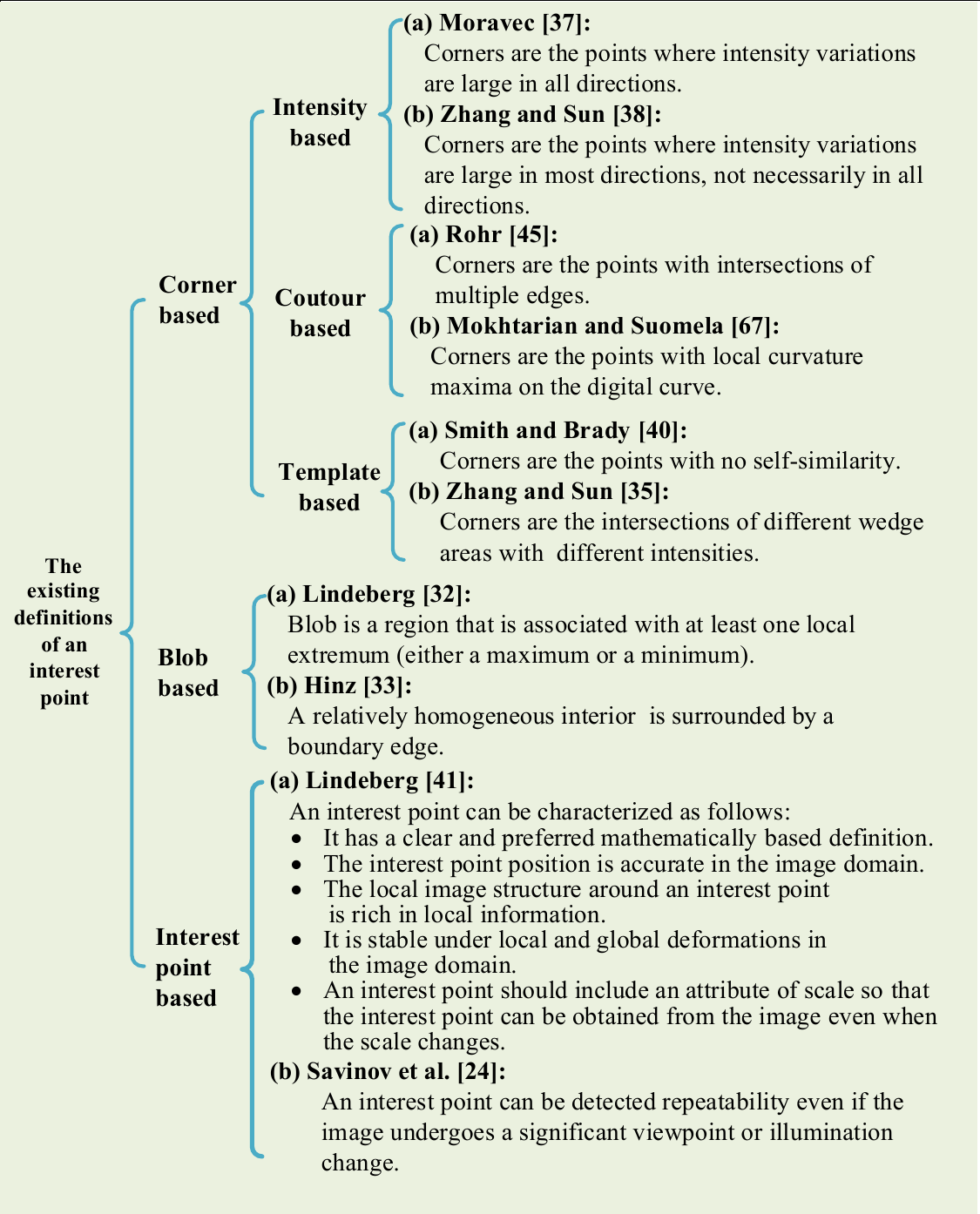}
\caption{Definitions of interest points.}
\label{fig2}
\end{figure}

It is indicated in~\cite{Mikolajczyk2002Indexing} that interest points include not only junctions, corners, and blobs, but also locations with significant texture variations. In fact, the terms `key point'~\cite{salti2013keypoints}, `dominant point'~\cite{nasser2018dominant}, `junction'~\cite{xia2014accurate}, `critical point'~\cite{zhu1995critical}, `corner'~\cite{harris1988combined}, `blob'~\cite{lindeberg1993detecting,hinz2005fast}, and `saliency feature'~\cite{banzhaf1998genetic} are special subclasses of `interest point'.
Up to now, scholars defined point features from the perspective of corners~\cite{zhang2019corner2,rohr1992recognizing}, blobs~\cite{lindeberg1993detecting}, and interest points~\cite{mikolajczyk2004scale}, and these definitions are shown in Fig.~\ref{fig2}. Furthermore, the existing classifications of corners fall into three categories: intensity-based~\cite{moravec1979visual, zhang2020corner}, contour-based~\cite{attneave1954some, zhang2019discrete}, and template-based~\cite{smith1997susan, zhang2019corner2}. As illustrated in Fig.~\ref{fig2}, the existing definitions of interest points include fuzzy concepts and precise mathematical definitions. There are contradictions between some definitions. For example, Moravec~\cite{moravec1979visual} defined that corners are the points where the intensity variations are large in all orientations. However, Zhang and Sun~\cite{zhang2020corner} defined that corners are the points where the intensity variations are large in most orientations, not necessarily in all orientations. Following these different definitions, different types of interest point detection methods extract IFI based on what they need from images and determine the interest points according to the extracted IFI.

Meanwhile, our investigation indicates that Lindeberg's definition~\cite{Lindeberg2015Image} on interest points is relatively comprehensive and has the following properties: (1) It has a clear and preferred mathematically based definition. (2) The interest point position is accurate in the image domain. (3) The local image structure around an interest point is rich in local information. (4) It is stable under local and global deformations in the image domain. (5) An interest point should include an attribute of scale so that the interest point can be obtained from the image even when the scale changes.

\begin{table*}[!htbp]
\setlength{\abovecaptionskip}{5pt}
\setlength{\belowcaptionskip}{0pt}
\footnotesize
\centering
\renewcommand{\arraystretch}{1.4}
\caption{Summary of related interest point detection surveys.}
\label{stest}
\begin{tabular}{llll}

\hline
Year& References & Venue &Content\\
\midrule[1pt]
1982& Kitchen and      & Pattern Recognition Letters   & A survey on contour based  corner detection   \\
~ &Rosenfeld~\cite{kitchen1982gray}&&algorithm till 1982.\\
1991& Deriche and     &IEEE Computer Vision and Pattern    & A survey on corner and vertex detectors. \\
~    &Giraudon~\cite{deriche1991corner}& Recognition&\\
1993& Abe et al.~\cite{abe1993comparison}     &IEEE International Conference on     & A survey and evaluation on corner detectors.  \\
~&~&Document Analysis and Recognition &\\
1994& Rohr~\cite{rohr1994localization}     & Journal of Mathematical Imaging and   & A survey on corner detectors. \\
~&~&Vision&\\
1996& Heyden and Rohr~\cite{heyden1996evaluation}     &IEEE International Conference on Pattern   & A survey and evaluation on five corner detectors   \\
~    &~& Recognition  &till 1996.\\
2000& Schmid et al.~\cite{schmid2000evaluation}     & International Journal of Computer Vision     & An evaluation on interest point detectors.\\
2001& Mohanna et al.~\cite{mohanna2001performance}     & British Machine Vision Conference  & An evaluation on five corner detectors under    \\
~&~&&image similarity and affine transforms.\\
2003& Sebe et al.~\cite{sebe2003evaluation}     & Image and Vision Computing  & An evaluation on salient point detectors. \\
2003& Rockett et al.~\cite{rockett2003performance}    &IEEE Transactions on Image Processing  & An evaluation on localization of corner detection \\
~&~& & methods.\\
2005& Mikolajczyk et al.~\cite{mikolajczyk2005comparison}     & International Journal of Computer Vision & An evaluation on blob detectors.\\
2005& Fraundorfer et al.~\cite{fraundorfer2005novel}     &IEEE Computer Vision and Pattern  & An evaluation on interest point detectors for\\
~    &~& Recognition& 3D scenes.\\
2006& Mokhtarian et al.~\cite{mokhtarian2006performance}     & Computer Vision and Image Understanding   & An evaluation on corner detectors.\\

2007& Moreels et al.~\cite{moreels2007evaluation}     & International Journal of Computer Vision  &An evaluation on feature detectors for 3D objects.  \\
2008& Tuytelaars et al.~\cite{tuytelaars2008local}   &  Foundations and Trends in Computer   & A broad survey on feature detectors.  \\
~&~&Graphics and Vision&\\
2010& Gil et al.~\cite{gil2010comparative}    & Machine Vision and Applications  & An evaluation on interest point detection methods.  \\
2010& Rosten et al.~\cite{rosten2008faster}     & IEEE Transactions on Pattern Analysis      & A high level summary on corner detectors before  \\
~&~&and Machine Intelligence&2010.\\
2011& Gauglitz et al.~\cite{gauglitz2011evaluation}     & International Journal of Computer Vision  & An evaluation on interest point detection methods  \\
~    &~& & in object tracking.  \\
2012& Aanaes et al.~\cite{aanaes2012interesting}     & International Journal of Computer Vision    &A thorough and detailed evaluation on feature   \\
~&~&~&detectors in an extensive dataset.\\
2012& Awrangjeb et al.~\cite{awrangjeb2012performance}     &IEEE Transactions on Image Processing   &An evaluation on contour-based corner detectors.\\
2012& Miksik et al.~\cite{miksik2012evaluation}     &IEEE International Conference on Pattern&An evaluation on feature detectors in image     \\
~    &~& Recognition  &matching.\\
2012& Heinly et al.~\cite{heinly2012comparative}     & European Conference on Computer Vision   & An evaluation on binary feature detectors since 2005.   \\
2013& Jiang et al.~\cite{jiang2013performance}     & Neurocomputing & An evaluation on interest point detectors based on   \\
~&~& &stereo visual odometry.\\
2013& Tombari et al.~\cite{tombari2013performance}     & International Journal of Computer Vision  &An evaluation on 3D keypoint detectors.   \\
2015& Li et al.~\cite{li2015survey}    & Neurocomputing  &A summary on feature detectors since 2000.  \\
2015 & Mukherjee et al.~\cite{mukherjee2015comparative}    & Machine Vision and Applications  &A summary on interest point detectors only    \\
~    &~&&covering recent milestones.\\
2015 &Rey et al.~\cite{rey2015comparing} &IEEE International Conference on Image  &An evaluation on feature detectors. \\
~    &~& Processing &\\
2016 &Henderson and  &Pattern Recognition Letters&An evaluation on feature bilateral symmetry.  \\
~    &Izquierdo~\cite{henderson2016symmetric}&&\\
2018& Komorowski et al.~\cite{komorowski2018interest}    & European Conference on Computer Vision   &An evaluation on learning based detectors based on   \\
~&~&& ApolloScape dataset. \\
2018& Lenc et al.~\cite{Lenc2018LargeSE}     & British Machine Vision Conference  &An evaluation on feature detectors based on homography  \\
~&~& &datasets.\\
2019 &Fan et al.~\cite{fan2019performance}     &IEEE  Transactions on Image Processing  &An evaluation on feature detectors for image-based  \\
~&~&~&3D reconstruction.\\
2020       &Rondao et al.~\cite{rondao2020benchmarking}     & Acta Astronautica  &An evaluation on eight detectors and descriptors.  \\
\hline
\end{tabular}
\end{table*}
\subsection{Main challenges}
\begin{figure}[!t]
\setlength{\abovecaptionskip}{2pt}
\setlength{\belowcaptionskip}{0pt}
\centering
\includegraphics[width=3.5in]{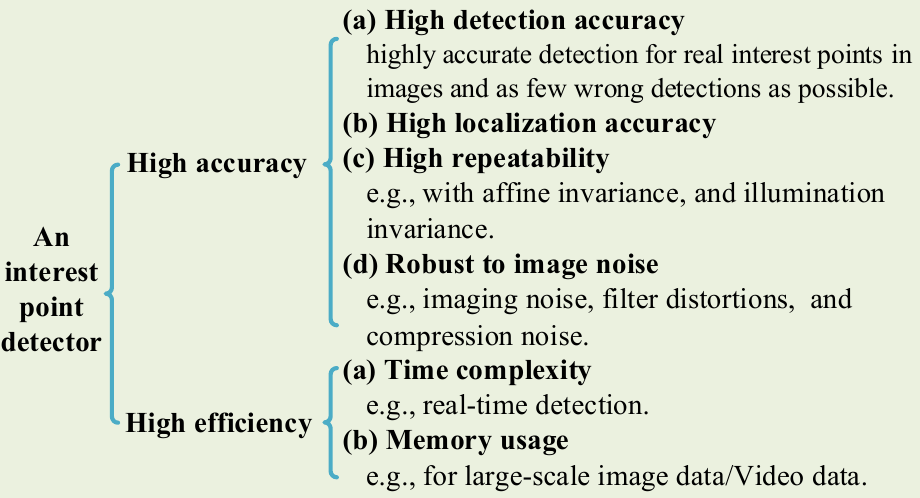}
\caption{Main challenges for interest point detection.}
\label{fig3}
\end{figure}

The ultimate goal of image interest point detection is to develop a general method that achieves of both high accuracy/quality and high efficiency as shown in Fig.~\ref{fig3}.

\begin{figure*}[!htbp]
\setlength{\abovecaptionskip}{2pt}
\setlength{\belowcaptionskip}{2pt}
\centering
\includegraphics[width=7in]{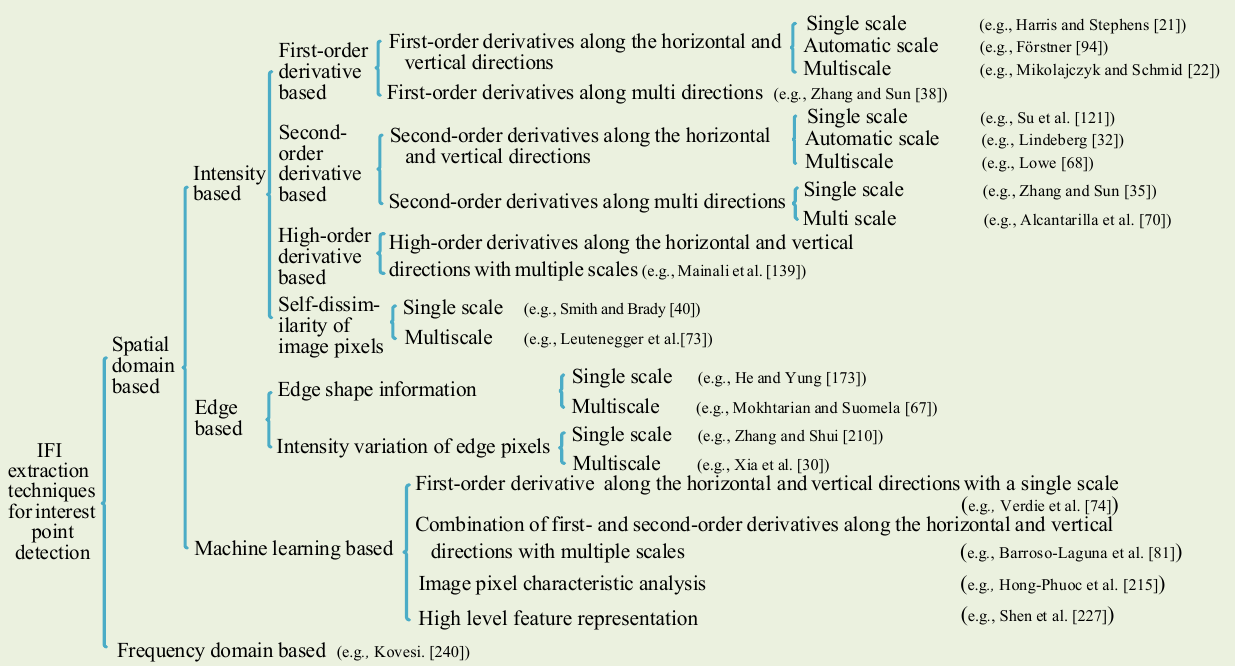}
\caption{Classification of the IFI extraction techniques in the existing interest point detection methods.}
\label{LSI}
\end{figure*}

On the one hand, high accuracy/quality interest point detectors should have the ability to accurately localize and extract interest points from images. It is well known that the reason why image interest point detection has been a research hot topic is that interest point detection has a wide range of applications in the field of image processing and computer vision. Most interest point based computer vision tasks rely on interest point detectors for finding local feature or interest point correspondences between different images such as for object tracking~\cite{lowry2015visual} and 3D reconstruction~\cite{fan2019performance}. For this purpose, high accuracy/quality interest point detectors should have high repeatability under different imaging conditions such as lighting directions, illuminations, camera poses, low resolution, blur, and different weathers. Other challenges for interest point detection come from digital artifacts, noise damage, poor resolution, and filtering distortion. On the other hand, high accuracy/quality interest point detectors should have the ability to detect interest points in real-time with acceptable memory and storage demands. The reason is that real-time processing is an indispensable requirement in many interest point based computer vision tasks such as autonomous driving~\cite{huang2016underwater} and image-guided robotic-assisted surgery~\cite{mirota2011system, shademan2016plenoptic}.

Our investigation also indicates that more accurate extraction of IFI makes interest point detection algorithms more likely to achieve better accuracy/quality, but it increases the complexity of the algorithms. This is a dilemma that still cannot be reconciled. This issue will be discussed in detail later in the paper.

\section{Image feature information extraction techniques for interest point detection}
\begin{figure*}[!t]
\setlength{\abovecaptionskip}{5pt}
\setlength{\belowcaptionskip}{0pt}
\centering
\includegraphics[width=7in]{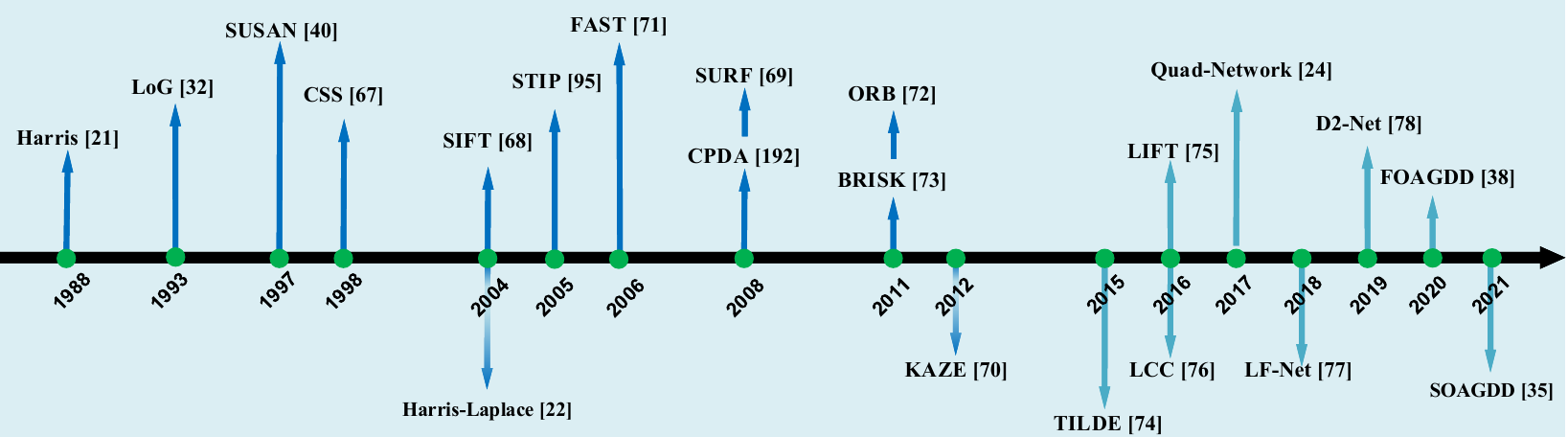}
\caption{Representative interest point detection methods. Starting from Harris method~\cite{harris1988combined}, first-order intensity variation based interest point detection methods gained remarkable popularity. In 1997, Smith and Brady~\cite{smith1997susan} proposed SUSAN method which marks the beginning of self-dissimilarity based interest point detection. In 1998, Mokhtarian and Suomela~\cite{mokhtarian1998robust} proposed CSS method which represents the edge contour based interest point detection methods received great popularity. In 1993, Lindeberg~\cite{lindeberg1993detecting} presented LoG method for detecting blobs from images in multi-scale space. Along this way, second-order intensity variation based blob detection methods (e.g., Harris-Laplace~\cite{mikolajczyk2004scale}, SIFT~\cite{lowe2004distinctive}, SURF~\cite{bay2006surf}, and KAZE~\cite{alcantarilla2012kaze}) were proposed for solving blob detection. In 2006, Rosten et al.~\cite{rosten2006machine} proposed FAST detector which is one of the first machine learning methods for interest point detection. Subsequently, ORB~\cite{rublee2011orb} and BRISK~\cite{leutenegger2011brisk} algorithm optimized the FAST detector~\cite{rosten2006machine}. In 2015, Verdie et al.~\cite{verdie2015tilde} proposed TILDE detector which is one of the first deep learning methods for interest point detection. And then, LIFT~\cite{yi2016lift}, LCC~\cite{lenc2016learning}, Quad-Network~\cite{savinov2017quad}, LF-Net~\cite{ono2018lf}, and D2-Net~\cite{dusmanu2019d2} were proposed by Yi et al., Lenc and Vedaldi, Savinov et al., Ono et al., and Dusmanu et al. respectively further promoting the development of deep learning based interest point detection. In 2020 and 2021, FOAGDD~\cite{zhang2020corner} and SOAGDD~\cite{zhang2019corner2} were proposed by Zhang and Sun which proved for the first time that first-order and second-order intensity variation along multiple directions have great help for improving the performance of interest point detection.}
\label{figtime}
\end{figure*}

In this section, we first introduce a taxonomy of the existing IFI extraction techniques for image interest point detection. According to this classification method, we discuss different types of IFI extraction techniques for interest point detection in depth. The representative interest point detection methods are shown in the Fig.~\ref{figtime}.

\subsection{A taxonomy on existing IFI extraction techniques for interest point detection}
The existing IFI extraction techniques for interest point extraction can be classified into two main categories: the spatial domain and the frequency domain based IFI extractions as illustrated in Fig.~\ref{LSI}. In addition, the spatial domain based IFI extraction techniques fall into intensity based, edge based, and machine learning based IFI extractions. Intensity based IFI extraction techniques extract image intensity variation information directly from images. Edge based IFI extraction techniques firstly extract edges from an input image by an edge detection algorithm (e.g., the Canny algorithm~\cite{Canny1986A} and the line segment detector~\cite{von2008lsd}), and then the shape or local structure information on edge pixels is obtained. The existing machine learning based IFI extraction techniques extract the first- or second-order derivatives information from images~\cite{verdie2015tilde, Barroso-Laguna2019ICCV}, or they learn high-level image abstractions from a designed network architecture. Frequency domain based IFI extraction techniques firstly transform the image from spatial domain to frequency domain, and then extract local structure information in the image in frequency domain.

There is a great degree of correlation between the aforementioned IFI extraction techniques given in Fig.~\ref{LSI} and the definitions of interest points shown in Fig.~\ref{fig2}. In the following, we describe in detail how different IFI extraction techniques extract corners, blobs, and interest points from images according to the classification of the IFI extraction techniques.

\subsection{Intensity based IFI extraction techniques for interest point detection}
The existing intensity based IFI extraction techniques can be divided into first-order derivative based, second-order derivative based, high-order derivative based, and self-dissimilarity on image pixels based approaches.

\subsubsection{First-order derivative based interest point detection}
Following Moravec's definition~\cite{moravec1979visual} that corners are the points with large variation in intensity in each direction, Harris and~Stephens~\cite{harris1988combined} constructed a $2$$\times$$2$ structure tensor for finding corners which have large variations on image intensities in horizontal and vertical orientations.  For a given 2D input image $I(x,y)$, the weighted sum of squared difference $\psi(u_x,u_y)$ is defined as
\begin{equation}\begin{aligned}
\label{eq1}
&\psi(u_x,u_y)=\int_{-\infty}^{\infty}\int_{-\infty}^{\infty}g_{\sigma}(x,y)\\
&~~~~~~~~~~~~~~\bigg(I(x+u_{x},y+u_{y})-I(x,y)\bigg)^{2}\text{d}x\text{d}y,
\end{aligned}\end{equation}
where $g_{\sigma}(x,y)$ is an isotropic Gaussian filter acting as the weighting function, $\sigma$ is a scale factor ($\sigma$$>$$0$), $(x,y)$ is a point location in the image, and $(u_x,u_y)$ is a local shift. The shifted image pixel $I(x+u_x,y+u_y)$ is approximated by a Taylor expansion truncated to the first order terms
\begin{equation}\begin{aligned}
\label{eq2}
I(x+u_{x},y+u_{y})\approx I(x,y)+u_{x}I_{x}(x,y)+u_{y}I_{y}(x,y),
\end{aligned}\end{equation}
where $I_{x}(x,y)$ and $I_{y}(x,y)$ denote the partial derivatives of the input image $\emph{I}$ with respect to the horizontal and vertical directions. Substituting approximation Equation~(\ref{eq2}) into Equation $(\ref{eq1})$ yields
\begin{equation}\begin{split}
\label{eq3}
\psi(u_{x},u_{y})\approx & \int_{-\infty}^{\infty}\int_{-\infty}^{\infty}g_{\sigma}(x,y)\\
&\bigg(u_{x}I_{x}(x,y)+u_{y}I_{y}(x,y)\bigg)^{2}\text{d}x\text{d}y\\
=&(u_{x}~u_{y})S{ u_{x} \choose u_{y}},
\end{split}\end{equation}
where $S$ is the structure tensor
\begin{equation}\begin{split}
\label{eq4}
S &= \int_{-\infty}^{\infty}\int_{-\infty}^{\infty}g_{\sigma}(x,y)\\
&\left[\begin{array}{cc}
I_{x}^2(x,y)&I_{x}(x,y)I_{y}(x,y)\\
I_{x}(x,y)I_{y}(x,y)&I_{y}^2(x,y)\\
\end{array}
\right]\text{d}x\text{d}y.\\
\end{split}\end{equation}
In~\cite{harris1988combined}, the attribute of each image pixel $(x,y)$ is determined by the eigenvalues $\lambda_1$ and $\lambda_2$ ($\lambda_1$$<$$\lambda_2$) of its corresponding structure tensor $S$. (1) If both $\lambda_1$ and $\lambda_2$ are close to zero, the pixel belongs to a flat region. (2) If $\lambda_1$ (or $\lambda_2$) is close to zero and $\lambda_2$ (or $\lambda_1$) has a large positive value, the pixel is an edge point. (3) If both $\lambda_1$ and $\lambda_2$ have large positive values, the pixel is a corner. To reduce the complexity of the algorithm, the determinant of the structure tensor $\text{Det}(S)$ and the trace of the structure tensor $\text{Trace}(S)$ are applied to construct the Harris corner measure as follows
\begin{equation}\begin{aligned}
\label{eq43}
\text{H}_a=\text{Det}(S)-k\times \text{Trace}^2(S),
\end{aligned}\end{equation}
where $k$ is a constant (e.g.,~$k$$=$0.04).

Following the Harris method, many different filters (e.g., Gaussian filter~\cite{shi1994good}, anisotropic nonlinear diffusion~\cite{brox2006nonlinear}, and Box filter~\cite{mainali2011robust}) with a single scale were used to extract image intensity variation information along a pair of orthogonal directions for detecting corners~\cite{shi1994good,zheng1999analysis, ando2000image, brox2006nonlinear, coleman2007concurrent, bastanlar2008corner, andreopoulos2008incremental, gevrekci2009illumination, cui2009automatic, loog2010improbability, mainali2011robust}. indicated that it is better to utilize the small eigenvalue of the structure tensor $S$ as the corner measure for detecting corners from an input image. Zheng et al.~\cite{zheng1999analysis} presented an equation with an adaptive parameter to replace the constant for the false corner response suppression in the Harris method for detecting corners. Ando~\cite{ando2000image} employed the first order derivatives along the horizontal and vertical directions to construct a gradient covariance matrix and the properties of the matrix are analyzed for detecting corners. For avoiding the influence of Gaussian image smoothing on the location of a feature, Brox et al.~\cite{brox2006nonlinear} replaced the isotropic Gaussian filter in the Harris method by an anisotropic nonlinear diffusion for extracting interest points. Similar to the Harris method, Coleman et~al.~\cite{coleman2007concurrent} used an isotropic Gaussian filter to smooth the input image and presented a corner measure based on the first-order derivatives along the horizontal and vertical directions. Bastanlar and Yardimci~\cite{bastanlar2008corner} improved the corner detection accuracy of the Harris method by estimating the geometry of the candidate corners. Andreopoulos et al.~\cite{andreopoulos2008incremental} inherited the Harris method, and the first-order derivatives along a pair of perpendicular directions are used to detect salient points in a series of incremental refinement images. Gevrekci and Gunturk~\cite{gevrekci2009illumination} presented an illumination robust corner detector by extending the Harris method to detect corners in a contrast-stretched image. Cui and Ngan~\cite{cui2009automatic} applied an angle range with a scale normalized Laplace of Gaussian (LoG) filter to detect interest points. Loog and Lauze~\cite{loog2010improbability} combined the corner measure of the Harris method and visual salience theory~\cite{gao2008plausibility} for detecting interest points from an input image. Mainali et al.~\cite{mainali2011robust} replaced the Gaussian filter in the Harris method with a Box filter to decrease the algorithm complexity of the Harris method.

Meanwhile, many algorithms extended the framework of the Harris algorithm from a single scale to an adaptive scale or multi-scales for improving the accuracy performance of corner detection. F{\"o}rstner et al.~\cite{forstner1994framework} applied an adaptive Wiener filter to smooth the input image and the first-order derivatives along the vertical and horizontal directions are used for detecting corners in the framework of the Harris algorithm. Laptev and Lindeberg~\cite{laptev2005spacerr} extended the Harris method in the spatial domain into the space-temporal domain, and the first-order orthogonal directional derivatives in the spatial domain and temporal derivative in the time domain are used to construct a space-time second-moment matrix for detecting interest points in the scale space in video. Mikolajczyk and Schmid~\cite{mikolajczyk2004scale} extended the single-scale Harris detector to a corner detector with adaptive scale selection. Furthermore, they proposed a scale invariant Harris-Laplace interest point detection method by combining the Harris corner detection method~\cite{harris1988combined} with the normalized Laplace operator~\cite{lindeberg1998feature}.  Gao et al.~\cite{gao2007multiscale} argued that it is appropriate to extract corners by using filters that are consistent with the human visual system. Then the log-Gabor wavelets with multiple scales were employed to extract multi-scale first-order orthogonal directional derivatives from an input image and construct multi-scale 2$\times$2 structure tensors for detecting corners. In view of the shearlets having the ability to efficiently obtain the anisotropic intensity variation information in images, Duval-Poo et al.~\cite{duval2015edges} applied the multi-scale shearlet filter to obtain IFI along a pair of orthogonal directions for detecting corners.

In addition, first-order orthogonal directional derivatives based IFI extraction techniques are often extended for detecting interest points in color or other type of images. Montesinos~\cite{montesinos1998differential} extended the Harris method, and the first-order derivatives along the horizontal and vertical directions are obtained in the RGB space to construct a 2$\times$2 structure tensor for detecting interest points in color images. Achard et~al.~\cite{achard2000sub} applied the Harris method for detecting corners in multispectral images. Weijer et~al.~\cite{van2005edge} extended the photometric invariance theory~\cite{geusebroek2001color} and presented a novel class of derivatives which is denoted as photometric quasi-invariants for improving the robustness of the obtained feature information. Then a corner detection method was proposed for color images in the framework of the Harris method. Ruzon and Tomasi~\cite{Ruzon2001Edge} utilized the earth mover's distance (EMD) technique~\cite{rubner1998metric} to identify the discrepancies of the color distributions for improving the accuracy of image corner detection. Weijer et~al.~\cite{van2005boosting} presented a salient point detector based on the distinctiveness of the local color information. The first-order orthogonal directional derivatives of the color image are used to measure the color distinctiveness~\cite{schmid2000evaluation} of a pixel. If the distinctiveness of the color derivatives of the pixel in a given neighborhood is the local maximum and larger than a given threshold, the pixel is marked as a salient point. Inspired by the optical flow technique, a pixel can be analyzed with arbitrary dimensions. Kenney et~al.~\cite{kenney2005axiomatic} combined the Harris method and an optical flow estimation technique for detecting corners in multi-channel images. In order to improve the robustness of corner detection in different imaging conditions, St{\"o}ttinger et~al.~\cite{stottinger2012sparse} extended the Harris method and the multi-scale image intensity variation information along a pair of orthogonal directions is used for selecting light invariant corners in an arbitrary color space.

Zhang and Sun~\cite{zhang2020corner} proved that the aforementioned interest point detection methods based on image intensity variation information along a pair of orthogonal directions cannot accurately extract interest points. It was indicated in~\cite{zhang2020corner} that first-order anisotropic Gaussian directional derivative (FOAGDD) filters have the high potential to accurately depict the differences between edges and corners. And then the multi-scale FOAGDDs of the image are used to construct a multi-directional structure tensor at multiple scales for detecting corners. Wang et~al.~\cite{wang2020corner} followed the FOAGDD method~\cite{zhang2020corner} for corner detection. In their method, an input image is smoothed by the multi-directional shearlet filters with multiple scales and a multi-directional structure tensor at multiple scales are constructed for detecting corners.

\begin{figure}[!htbp]
\setlength{\abovecaptionskip}{5pt}
\setlength{\belowcaptionskip}{0pt}
\centering
\includegraphics[width=3.5in]{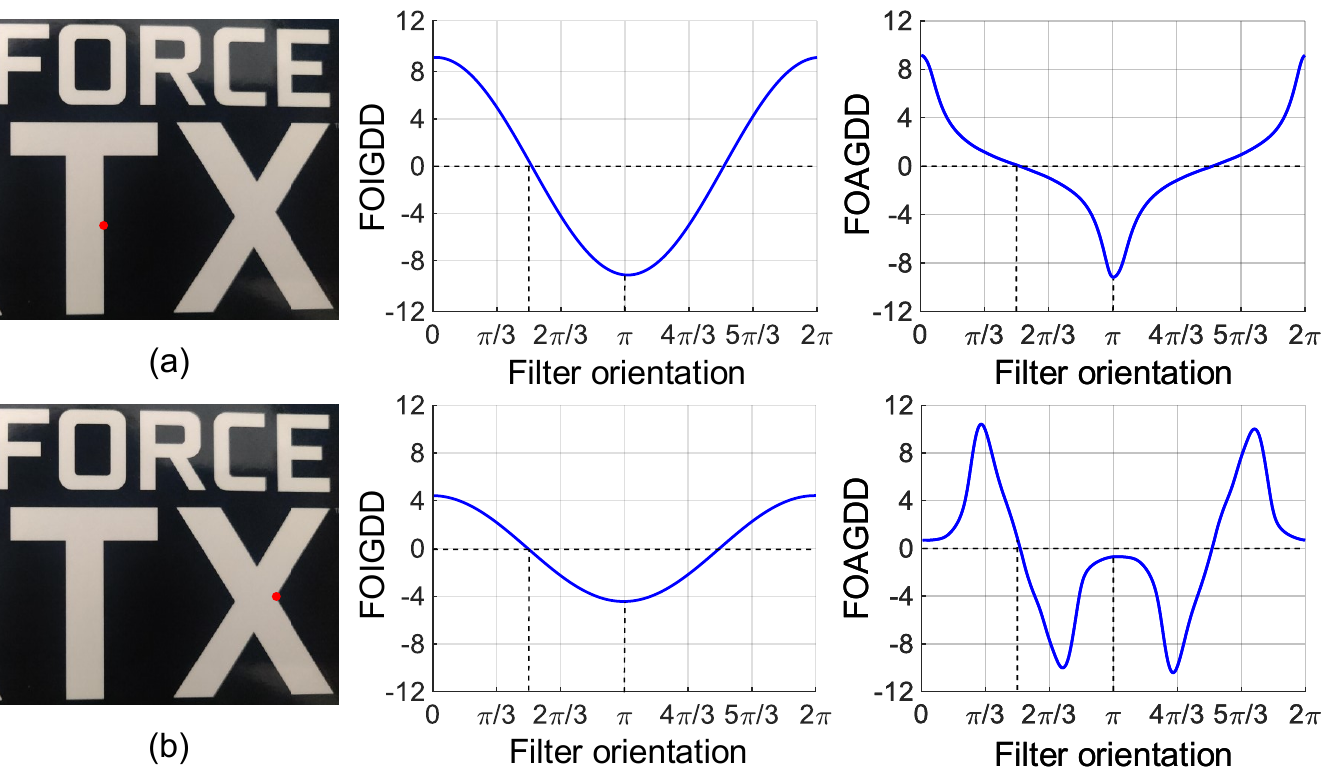}
\caption{A step edge point and an L-type corner are shown in (a) and (b) respectively (the red dots). Their corresponding first-order isotropic Gaussian directional derivative (FOIGDD) and first-order anisotropic Gaussian directional derivative (FOAGDD) are shown in the second and third columns respectively.}
\label{fig5}
\end{figure}

Zhang and Sun~\cite{zhang2020corner} indicated that the first-order derivatives along the horizontal and vertical directions do not have the ability to accurately extract IFI from an input image and do not have the ability to accurately depict the differences between edges and corners. The reasons are as follows~\cite{zhang2020corner}: firstly, the intensity variation around a corner is large in most directions, not necessarily in all directions. The first-order derivatives along the horizontal and vertical directions cannot properly extract local intensity variation information for detecting corners. Secondly, the first-order derivatives along the horizontal and vertical directions cannot depict the intensity variations at step edges and corners well. Thirdly, the isotropic Gaussian filter cannot accurately depict the intensity variation differences between step edges and corners. As shown in Fig.~\ref{fig5}(b), for the L-type corner, it is obvious from the second and third columns of Fig.~\ref{fig5} that the directional derivatives are large in most directions at a corner while their directional derivatives of the isotropic or the anisotropic Gaussian filters are very small or even near zero along the vertical ($0$) or horizontal ($\pi/2$) directions. Then, this type of corners may not be correctly detected by the Harris detector. This phenomenon does not satisfy the definition for a corner~\cite{moravec1979visual} that the directional derivatives are large in all directions. Furthermore, the directional derivatives along the horizontal and vertical directions cannot accurately depict the intensity variation differences between edges and corners. Take a step edge and an L-type corner as examples as shown in the first column of Fig.~\ref{fig5}, their corresponding directional derivatives are zero in the horizontal direction. However, in the vertical direction, the absolute magnitude of the directional derivative on the edge is larger than that at the L-type corner. Then, an edge pixel may be detected as a corner by Equation~(\ref{eq4}), while a real corner may be marked as an edge. The reason is that all the isotropic Gaussian derivatives along 0 to 2$\pi$ filtering direction of a step edge, L-, Y-, X-, and star-type corners are proved~\cite{zhang2020corner} to satisfy with the waves of the sine functions.

\begin{figure}[h!]
\setlength{\abovecaptionskip}{5pt}
\setlength{\belowcaptionskip}{0pt}
\centering
\includegraphics[width=3.5in]{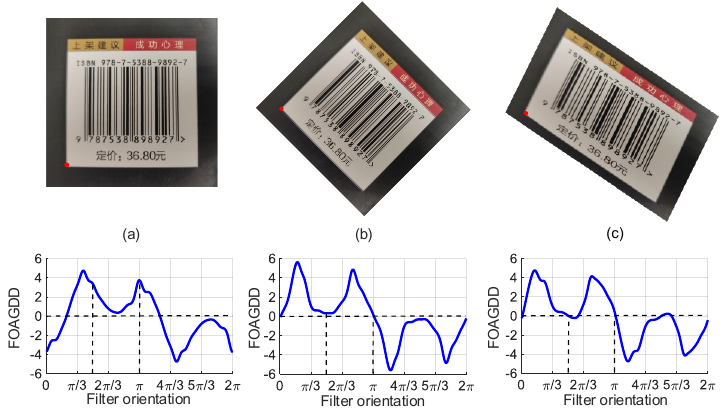}
\caption{Examples of first-order anisotropic Gaussian directional derivative changes by image rotation or affine image transformation. (a)~an L-type corner (shown as the red dot), (b)~the L-type corner is rotated by $\pi/4$ clockwise, (c)~the L-type corner undergoes an affine image transformation.}
\label{fig6}
\end{figure}

Furthermore, Zhang and Sun~\cite{zhang2020corner} indicated that two orthogonal first-order anisotropic Gaussian directional derivatives along the horizontal and vertical directions cannot accurately detect corners on an affine transformed image. Take an L-type corner as an example as shown in Fig.~\ref{fig6}(a) (the red dots), its corresponding two orthogonal anisotropic Gaussian directional derivatives are large as shown in the second column of Fig.~\ref{fig6}. According to the criteria of Harris corner detection, it can be detected as a corner. After the L-type corner is rotated by $\pi/4$ clockwise as shown in Fig.~\ref{fig6}(b), its corresponding two orthogonal directional derivatives are small as shown in the second column of Fig.~\ref{fig6}. Then the corner may not be detected with such or similar image rotation transformations. The reason is that the two orthogonal directional derivatives do not contain enough local structure information. It is worth to note that multi-scale filtering techniques along the horizontal and vertical directions~\cite{mikolajczyk2004scale} cannot solve the aforementioned problem because the multi-scale filtering technique along the horizontal and vertical directions only efficiently enhance the local intensity variation extraction along the horizontal and vertical directions. Another example is when the image is rotated and squeezed, which means that the shape of the corner is changed. If the L-type corner undergoes an affine image transformation as shown in Fig.~\ref{fig6}(c), its corresponding two orthogonal directional derivatives are also small as shown in the second column of Fig.~\ref{fig6}. Then the corner may not be detected with such or similar affine image transformations. The multi-scale filtering techniques along the horizontal and vertical directions cannot solve the aforementioned problem either.

It is worth to note that gradient amplitude or gradient direction information can also be used to detect interest points with high local symmetry/mirror regions in images, because the center of a circular blob feature has a high degree of local mirror symmetry. Reisfeld et~al.~\cite{reisfeld1995context} applied the concept of symmetry to compute a symmetry map from an input image. For each pixel, the symmetry strength is calculated by looking at the magnitude and the direction of the derivatives of neighboring pixels. Pixels with high symmetry are marked as interest points. Heidemann~\cite{heidemann2004focus} extended the method in~\cite{reisfeld1995context} and interest points are detected in color images. Loy and Zelinsky~\cite{loy2003fast} proposed a fast interest point detection algorithm and used the information on the gradient direction distribution of pixels to detect interest points in local radial symmetric regions in the image. Maver~\cite{maver2010self} exploited the concept of self-similarity to detect interest points associated with symmetrical regions in images in scale space.

\subsubsection{Second-order derivative based interest point detection}
In~\cite{beaudet1978}, the second-order horizontal, vertical, and cross derivatives ($\hbar_{x x}(x,y)$, $\hbar_{y y}(x,y)$, and $\hbar_{x y}(x,y)$) of $\hbar(x,y)$ are derived to construct an Hessian matrix $\mathbf{H}$
\begin{equation}
\label{eq6}
\mathbf{H}=\left[\begin{array}{ll}{\hbar_{x x}} & {\hbar_{x y}} \\ {\hbar_{x y}} & {\hbar_{y y}}\end{array}\right].
\end{equation}
Corners are defined as the local extremum (either a maximum or a minimum) in the determinant of the Hessian
\begin{equation}
\label{eq7}
C=|\mathbf{H}(\hbar)|=\hbar_{x x}\hbar_{y y}-(\hbar_{x y})^2.
\end{equation}

Following the Beaudet method~\cite{beaudet1978}, second-order derivative based IFI extraction techniques were widely used for interest point detection.

Lindeberg~\cite{lindeberg1993detecting} defined a blob as a region associated with at least one local extremum (either a maximum or a minimum) and then a multi-scale Laplacian of Gaussian (LoG) filter is used to extract the second-order horizontal and vertical directional derivatives from an input image for detecting blobs.  Wang and Brady~\cite{wang1995real} considered an input image as a surface, and the first- and second-order derivatives of an input image along the horizontal and vertical directions are obtained to look for corners where there are large curvature changes along an edge. In~\cite{lindeberg1998feature}, Lindeberg utilized the normalized Laplace operator with adaptive scales for detecting corners. Abdeljaoued and Ebrahimi~\cite{abdeljaoued2004feature} followed the scale-space theory~\cite{inbook} to extract interest points as follows
\begin{equation}\begin{aligned}
\label{eq8}
K_{\text{norm}}(x,y,\sigma)=\sigma^2(\hbar_{x x}\hbar_{y}^2+\hbar_{y y}\hbar_{y}^2-2\hbar_{x y}\hbar_{x}\hbar_{y}).
\end{aligned}\end{equation}
where $\hbar_{x}$, and $\hbar_{y}$ denote the partial derivative of the $\hbar(x,y)$ with respect to the horizontal and vertical directions. In order to reduce computational complexity of LoG~\cite{lindeberg1993detecting} and efficiently construct a scale space pyramid, Lowe~\cite{lowe2004distinctive} proposed the scale invariant feature transform (SIFT) method which approximated the normalized LoG filter by a difference of Gaussian (DoG) filter
\begin{equation}\label{eq9}
D(x, y, \sigma)=(g(x, y, q \sigma)-g(x, y, \sigma)) \otimes I(x, y),
\end{equation}
where $q$ is a constant. The candidate interest points are detected by seeking for local maxima in a DoG pyramid. Meanwhile, the eigenvalues ($\lambda_{\text{max}}$, $\lambda_{\text{min}}$, with $\lambda_{\text{max}}$$>$$\lambda_{\text{min}}$) of the Hessian matrix (the second-order derivative extracted by DoG) are applied to suppress edge responses
\begin{equation}\begin{aligned}
\left [ \begin{array}{cc}
{D_{x x}}&{D_{x y}}\\
{D_{x y}}&{D_{y y}}
\end{array} \right],\nonumber\end{aligned}\end{equation}
where $D_{x x}$ and $D_{y y}$ denote the second-order orthogonal directional derivatives of $D(x,y)$, and $D_{x y}$ denotes the second-order cross derivative of $D(x,y)$. For each candidate interest point, if the ratio of the two eigenvalues ($\lambda_{\text{max}}$ and $\lambda_{\text{min}}$) is larger than $\varsigma$, i.e.,
\begin{equation}\begin{aligned}
\label{eq11}
~~~~~~~~~~~~\frac{\lambda_{\text{max}}}{\lambda_{\text{min}}}>\varsigma,
\end{aligned}\end{equation}
the candidate interest point may be marked as an edge point (e.g.,~$\varsigma$$=$$6$).

Following the LoG~\cite{lindeberg1993detecting} and DoG~\cite{lowe2004distinctive} methods, different types of filters (e.g., piece-wise triangle filters~\cite{marimon2010darts} or Box filters~\cite{mcdonnell1981box, gao2020fast}) were used to obtain the second-order derivatives along a pair of orthogonal directions with multiple scales or adaptive scales from an input image for detecting interest points. Bay et~al.~\cite{bay2006surf} aimed to reduce the calculation complexity in image convolution of the DoG method~\cite{lowe2004distinctive}. Box filters~\cite{mcdonnell1981box} and integral images are utilized to obtain a similar second-order horizontal and vertical directional derivatives for detecting interest points from images in scale space. Agrawal et~al.~\cite{agrawal2008censure} approximated the DoG filters~\cite{lowe2004distinctive} with the centers-surround filters for detecting features in the framework of the DoG method. To improve the accuracy and repeatability of interest points detection, Marimon et~al.~\cite{marimon2010darts} applied piece-wise triangle filters to approximate the second-order Gaussian derivatives along a pair of orthogonal directions with multiple scales for searching extrema in the scale space. Su et~al.~\cite{su2012junction} applied the second-order Gaussian horizontal and vertical directional derivatives to construct a Hessian matrix for detecting junctions in blood vessel images.
\indent Alcantarilla et~al.~\cite{alcantarilla2012kaze} indicated that a Gaussian filter with the same degree of smoothing on details and noise in images reduces the localization accuracy of detected interest points. And they proposed a KAZE interest point detector. Additive operator splitting schemes~\cite{weickert1998efficient} are used to approximate the Perona and Malik diffusion equation~\cite{perona1990scale} and the local maxima of the Hessian in a nonlinear scale space are used for detecting interest points. Later, an accelerated KAZE method~\cite{alcantarilla2011fast} was presented to reduce the computation complexity of the KAZE method~\cite{alcantarilla2012kaze}. In~\cite{mainali2013sifer}, Mainali et~al. proved that the designed cosine modulated Gaussian filter has a balanced scale-space atom with minimal spread in both scale and space. Then the designed filter is employed to obtain the second-order horizontal and vertical direction derivatives from an input image for detecting interest points with an adaptive scale in the framework of the LoG method~\cite{lindeberg1993detecting}. Furthermore, edges are eliminated using the condition given in~(\ref{eq11}) from detected interest points. To improve the robustness of interest point detection in abrupt variations of images caused by illumination or geometric transformation, Miao and Jiang~\cite{miao2013interest} utilized the weighted rank order filter to replace the LoG filter to detect interest points in the framework of the LoG method. In~\cite{salti2013keypoints}, Salti et~al. demonstrated that the multi-scale wave equation~\cite{mainardi1996fundamental} can be effectively used to detect keypoints with significant symmetries at different scales. Then the multi-scale wave equation~\cite{mainardi1996fundamental} was utilized to obtain the second-order derivatives from an input image for finding local extrema points as keypoints in the scale space. Li et~al.~\cite{li2014ga} utilized the geometric algebra theory~\cite{hestenes2012clifford} to detect interest points in multispectral images in the framework of the DoG method. In order to improve the accuracy of interest point detection in low contrast images, Miao et~al.~\cite{miao2015contrast} utilized the zero-norm LoG filter for detecting interest points from images. Xu et~al.~\cite{xu2016dfob} followed the CenSurE method~\cite{agrawal2008censure}, and an octagon filter bank was used to obtain the second-order derivatives along the horizontal and vertical directions for detecting blobs in the scale space. Because shearlet filters have good IFI extraction ability and noise robustness, Duval-Poo et~al.~\cite{duval2017scale} employed the multi-scale shearlet filters to detect blobs in the framework of the DoG method. Marsh et~al.~\cite{marsh2018hessian} applied the LoG filters and the determinant of the Hessian to locate blobs in atomic force microscopy images. Wang et~al.~\cite{wang2019ga} combined the SURF method~\cite{bay2006surf} with the geometric algebra theory~\cite{hestenes2012clifford} for detecting interest points in multispectral images.

Alternately, multi-directional second-order derivatives with multiple scales were applied to detect interest points from an input image. Kong et~al.~\cite{kong2013generalized} applied multi-directional generalized (isotropic and anisotropic) LoG filters to smooth the input image in scale space, and then the amplitude responses in different directions are multiplied by an empirical coefficient; and the sum of the amplitude responses in different directions are used to detect blobs in the scale space. Li and Shui~\cite{li2020subpixel} followed the Kong et~al.~\cite{kong2013generalized} method for extracting blobs from an input image, and edge suppression using the condition in~(\ref{eq11}) is applied for edge points removal from detected blobs. Wang et~al.~\cite{wang2020automated} presented a normalized method for the LoG filters~\cite{lindeberg1993detecting} and then the multi-directional normalized filters with multiple scales are used to obtain the multi-directional second-order derivatives from an input image, and then the smallest directional derivative in multiple directional derivatives is used to detect blobs in the scale space.Ghahremani et~al.~\cite{ghahremani2020ffd} introduced in detail how the DoG method~\cite{lowe2004distinctive} should set multi-scale parameters. Then a novel discrete multi-scale pyramid is designed using the proposed the scale-space pyramid's blurring ratio and smoothness for improving the accuracy of interest point detection.
\begin{figure}[!htbp]
\setlength{\abovecaptionskip}{5pt}
\setlength{\belowcaptionskip}{0pt}
\centering
\includegraphics[width=3.5in]{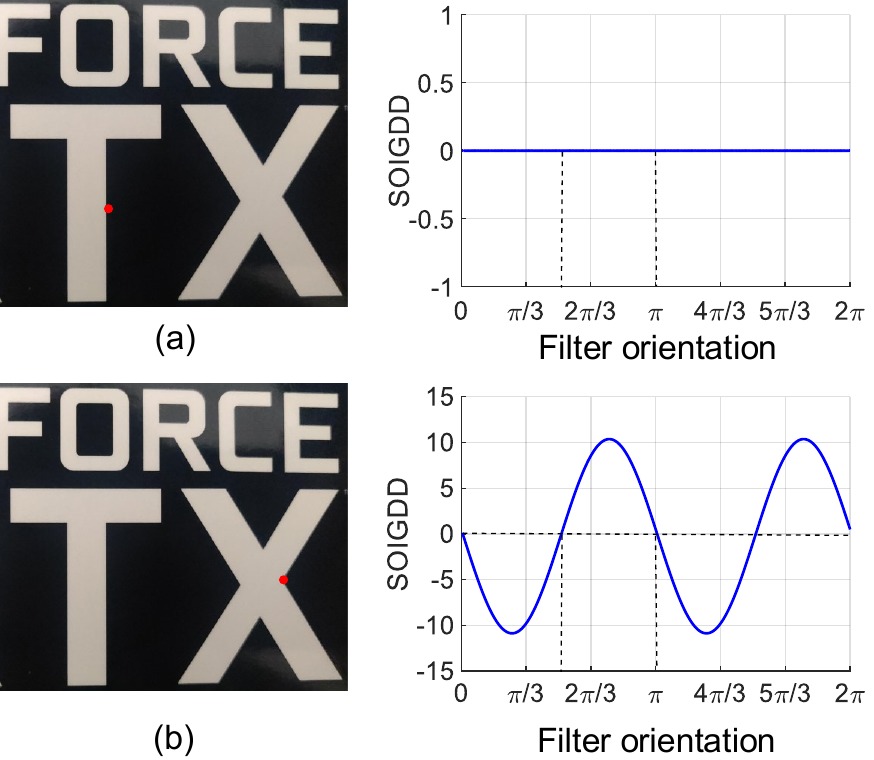}
\caption{A step edge and an L-type corner (the red dots) are shown in (a)-(b) in the first column. Their corresponding second-order isotropic Gaussian directional derivatives (SOIGDDs) are shown in the second column.}
\label{fig7}
\end{figure}
Zhang and Sun~\cite{zhang2019corner2} applied the second-order generalized Gaussian directional derivative (SOGGDD) filters to derive the SOGGDD representations of step edge, L-type corner, Y- or T-type corner, X-type corner, and star-type corner. In order to reduce the time complexity of the algorithms, Gao et al.~\cite{gao2020fast} applied integral image and box filter to approximate the convolution process of SOGGDD filter and proposed a coarse screening mechanism based on the sum of multi-directional second order derivatives to detect corner. The SOGGDD representations of a step edge and different types of corners indicate that the SOGGDD of a step edge is zero in each direction and the SOGGDD filters along a pair of orthogonal directions cannot obtain enough IFI to accurately describe the second-order directional derivative at corners. Take an L-type corner and a step edge as examples as illustrated in Fig.~\ref{fig7}, it can be seen that the second-order isotropic Gaussian directional derivative (SOIGDD) of the corner is very large in many directions and the SOIGDD of the step edge is zero in all orientations. Meanwhile, it can be seen from Fig.~\ref{fig7} that the SOIGDDs of the corner along the orthogonal directions are zero. It means that the ratio of the two eigenvalues of the 2$\times$2 Hessian matrix cannot be used to properly describe the attribute of an image pixel. It is indispensable to extract the second-order derivative with multiple directions for detecting corners.

\subsubsection{High order derivative based interest point detection}
Mainali et~al.~\cite{mainali2014derivative} proved that the 10th-order multi-scale Gaussian derivative filter reaches the jointly optimal Heisenberg's uncertainty principle~\cite{heisenberg1985anschaulichen} of its impulse response in scale and space simultaneously. Then, the 10th-order horizontal and vertical Gaussian directional derivatives are obtained from an input image for detecting interest points. Furthermore, a 2$\times$2 matrix based on the 10th-order derivatives is constructed which is similar to condition~(\ref{eq11}) for eliminating edge pixels from the detected interest points. In order to improve the accuracy of interest point detection when the input image is blurred, Saad and Hirakawa~\cite{Saad2016Defocus} extended the SIFT~\cite{lowe2004distinctive} method in which Gaussian filters with multiple scales are used to obtain the 4th-order Gaussian horizontal and vertical directional derivatives from an input image for detecting interest points in scale space.

\subsubsection{Self-dissimilarity of image pixels based interest point detection}
It is well-known that the apex of a wedge is not self-similar. The apex of the wedge corresponds to a corner point. Given the size of a circular mask, the similarity ratio information between the gray values at the pixels in the circular mask and the gray value at the center pixel of the template can be used to measure the self-similarity of the center pixel.

Smith and Brady~\cite{smith1997susan} used the similarity ratio information for detecting corners from an input image where a corner is defined as a point with smallest univalue segment assimilating nucleus (SUSAN).  Trajkovic and Hedley~\cite{Trajkovi1998Fast} extended the SUSAN method, and the corner measure of image pixel is defined as minimum intensity variations over all possible orientations. If the corner measure is higher than a given threshold, the image pixel is marked as a corner.  Self-dissimilarity of image pixels can also be measured by local entropy. Gilles~\cite{gilles1998robust} applied the local intensity histograms to calculate the local entropy within a given template for detecting interest points. Cazorla et~al.~\cite{cazorla2002junction} applied the SUSAN method to extract candidate corners, and then a probabilistic edge model and a log-likelihood test are combined for determining junction points. Sun et~al.~\cite{Sun2002COP} inherited the SUSAN method, and two oriented cross templates are used to calculate the self-dissimilarity information between the center pixel and the pixels around the center pixel in a given template for detecting corners. Cazorla and Escolano~\cite{Cazorla2003Two} applied the SUSAN method to find candidates and then intensity distributions are used to extract corners from the candidates. Following the SUSAN method~\cite{smith1997susan}, Rosten et~al.~\cite{rosten2006machine, rosten2008faster} presented the features from the accelerated segment test (FAST) algorithm. The pixel value differences between a target point and a circular neighborhood template are utilized for detecting corners from an input image. Mair et~al.~\cite{mair2010adaptive} improved the FAST method~\cite{rosten2008faster} by finding the optimal decision tree. Bennett and Lasenby~\cite{bennett2014chess} proposed a method for detecting corners on a chess-board. A ring of 16 pixels around each pixel in a chess-board image are sampled at a constant radius with equal angular spacing. The gray values of these 16 pixels are analyzed to determine whether the center pixel belongs to an edge, a corner, or a flat region. Hasegawa et~al.~\cite{hasegawa2014keypoint} improved the size of the template to calculate the self-dissimilarity for detecting corners in the framework of the FAST method. Bok et~al.~\cite{bok2016automated} utilized the special characteristics of corners on a chess-board (e.g., four black-white color changes on a circular boundary around a corner) to automatically localize corners, and gradient slopes are used to discard false corners.

It was indicated in~\cite{leutenegger2011brisk} that the FAST method does not account for scale variations, and a scale space modification of the FAST method is employed for detecting corners. Rublee et~al.~\cite{rublee2011orb} applied the multi-scale pyramidal representation~\cite{lowe2004distinctive} into the FAST method and corners are extracted in the scale space. Furthermore, the Harris method is employed for edge rejection. Buoncompagni et~al.~\cite{buoncompagni2015saliency} presented a key point extraction algorithm based on the FAST method and the BRIEF descriptor~\cite{calonder2010brief}. If candidate points extracted by the FAST method have high repeatability in target detection and image matching, these candidates are marked as key points; otherwise, they are deleted.

\subsection{Edge contour based IFI extraction techniques for interest point detection}
The existing edge contour based IFI extraction techniques fall into two categories: edge shape information and intensity variation on edge pixels.

\subsubsection{Edge shape based interest point detection}
In general, the existing edge shape-based interest point detection methods can be divided into two categories~\cite{Mohammad2012Performance}: direct curvature estimation methods and indirect curvature estimation methods.

Direct curvature estimation techniques detect corners by estimating the cosine of angle, local curvature, or tangential deflection at each point on edge contours. In the early developed algorithms, angular change was used to find corners~\cite{rosenfeld1973angle, Rosenfeld1975an, Beus1987an, teh1989detection} on edge contours. Alternately, isolated discontinuity points in the mean slope were used to detect corners~\cite{Freeman1977A, Beus1987an}. Asada and Brady~\cite{Asada1986The} introduced the scale space concept to represent significant changes in curvature along a planar curve for detecting corners. Rattarangsi and Chin~\cite{rattarangsi1990scale} analyzed the curvature characteristics of the L-, END-, and STAIR-models in the Gaussian scale space and developed a multiple-scale corner detection method. Lee et~al.~\cite{lee1993wavelet, lee1995multiscale} applied multi-scale wavelet
filters to smooth the edge contours and corners are evaluated by the angles between the edge pixels and their neighbouring edge pixels. Cooper et~al.~\cite{cooper1993early} introduced a corner measure which uses the pixel coordinates of chain-codes for estimating the curvature for detecting corners. Ferm{\"u}ller and Kropatsch~\cite{fermuller1994syntactic} applied a discrete pyramid technique~\cite{Kropatsch1985A} to analyze the curvature characteristics of edge pixels in scale space. Mokhtarian and Suomela~\cite{mokhtarian1998robust} presented a corner detection method with multi-scale based on the curvature scale-space (CSS) representations~\cite{mokhtarian1986scale, Mokhtarian1992A}. For a given parametric vector equation $\Gamma(l)=\{u(l), v(l)\}$ of a planar curve, the Gaussian filter with multiple scales $g(l, \sigma)$ are employed to smooth the planar curve. Then the multi-scale curvature $\kappa(l, \sigma)$ on the smoothed curve is
\begin{equation} \label{eq12}
\kappa(l, \sigma)=\frac{\dot{U}(l, \sigma) \ddot{V}(l, \sigma)-\ddot{U}(l, \sigma) \dot{V}(l, \sigma)}{\left(\dot{U}^{2}(l, \sigma)+\dot{V}^{2}(l, \sigma)\right)^{3 / 2}},
\end{equation}
with $\dot{U}(l, \sigma)=u(l) \otimes \dot{g}(l, \sigma)$, $\dot{V}(l, \sigma)=v(l) \otimes \dot{g}(l, \sigma)$, $\ddot{U}(l, \sigma)=u(l) \otimes \ddot{g}(l, \sigma)$, and $\ddot{V}(l, \sigma)=v(l) \otimes \ddot{g}(l, \sigma)$.  $\otimes$ refers to convolution, $\sigma$ is the scale factor, and $\dot{g}(l, \sigma)$ and $\ddot{g}(l, \sigma)$ are the first- and second-order derivatives of $g(l, \sigma)$.

Beau and Singer~\cite{beau2001reduced} applied a hierarchical discrete correlation algorithm~\cite{burt1981fast} for detecting corners on edge contours where corners on edge contours should always be detected in different scale spaces. Ray and Pandyanp~\cite{ray2003acord} replaced the multi-scale Gaussian smoothing~\cite{mokhtarian1998robust} by an adaptive smoothing on edge contours, and then the curvature in Equation~(\ref{eq12}) for each edge pixel is calculated for detecting corners. Inspired by~\cite{rattarangsi1990scale}, Zhong and Liao~\cite{zhong2007direct} presented a theoretical analysis on the direct curvature scale space (DCSS) for detecting corners on edge contours. In order to improve the peak value of a corner response while suppressing noise response through Gaussian smoothing, Zhang et~al.~\cite{zhang2007multi} extended the CSS method~\cite{mokhtarian1998robust} with the application of multi-scale curvature product of pixels for detecting corners. In order to eliminate the false corners in the corner set, He and Yung~\cite{he2008corner} extended the CSS method~\cite{mokhtarian1998robust} by applying an adaptive local threshold based on the mean curvature within a region of support. Nguyen and Debled-Rennesson~\cite{nguyen2011discrete} combined the discrete geometry theory~\cite{debled2006optimal} with curvature calculation~\cite{mokhtarian1998robust} to analyze the shape characteristics of edge contours for detecting corners. Pham et~al.~\cite{pham2014accurate} proposed an algorithm to extract corners by finding the points that are intersections of multiple edges. Zhang et~al.~\cite{zhang2015laplacian} replaced the multi-scale Gaussian filtering~\cite{rattarangsi1990scale} by Laplacian of Gaussian to analyze the behaviors of curvatures of the L-, END-, and STAIR-models in a multi-scale space for detecting corners. Chen et~al.~\cite{chen2016kd} extended the angle based curvature technique~\cite{borrelli2010error} from continuous domain to discrete domain for detecting corners. Lin et~al.~\cite{lin2017efficient} employed the second-order difference operator to depict the difference between corners and edge pixels on edge contours for detecting corners. Zhang et~al.~\cite{zhang2019discrete} identified that the existing curvature calculation~\cite{mokhtarian1998robust} is not robust to local change and noise in the discrete domain and has poor accuracy for corner detection when corners are closely located. And then discrete curvature representations of the L-, END-, and STAIR-models are derived and investigated for detecting corners. For improving the computation efficiency, Wang et al.~\cite{wang2021fast} applied difference operation to approximate the intensity derivative for calculating the contour curvature and defined the local maximum point as a corner point.

Meanwhile, indirect curvature estimation techniques have also been widely used for detecting corners on edge contours. Han et~al.~\cite{han1989identification} proposed a line splitting method to recursively analyze the characteristics of contours for detecting corners. Illing and Fairney~\cite{illing1991determining} firstly searched boundary curves for segments which exhibit significant linearity, and then they analyzed the curvature at the junctions between these boundary segments for detecting corners on edge contours. It is well known that corners are rich in high frequency components. Pei and Horng~\cite{pei1994corner} applied a low-pass filter to suppress corners and resulted in a smooth curve. Then they detected corners based on the fact that corners have a larger amount of shift than edge pixels. Zhu and Chirlian~\cite{zhu1995critical} indicated that the existing discrete curvature calculation methods have quantization errors. Therefore, shape representation (e.g., for a T-type corner), shape characteristic analysis, and shape recognition are used for detecting corners on edge contours. Ji and Haralick~\cite{ji1998breakpoint} proposed a statistical approach for detecting corners on edge contours. An edge pixel is declared as a corner if the estimated orientations of the two fitted lines of the two edge segments immediately to the right and left of the edge contour point are significantly different. Li and Chen~\cite{li1999corner} estimated a set of fuzzy patterns for contour points. And then corner detection is characterized as a fuzzy classification problem based on the edge contours. Neumann and Teisseron~\cite{neumann2002extraction} extended the Freeman code technique~\cite{Freeman1977A} to describe contours, and then the angular variation for each pixel in the region of support~\cite{held1994towards} is used to find corners. Awrangjeb and Lu~\cite{awrangjeb2008improved} applied the affine-length parameterization to compute the discrete curvature for detecting corners. Lagani{\`e}re~\cite{laganiere1998morphological} applied the mathematical morphology closing operator for detecting corners on edge contours. Sobania and Evans~\cite{sobania2005morphological} used a range of mathematical morphology operations for detecting corners on edge contours. According to the geometric characteristics of corners, a pair of differently sized and oriented triangular morphological structuring elements are utilized to match and locate corners on edge contours.In order to enhance the robustness for corner detection with image affine transformations, Awrangjeb and Lu~\cite{awrangjeb2008robust} applied the chord-to-point accumulation (CPDA) technique~\cite{Han2001Chord} for calculating the discrete curvature and detecting corners on edge contours. Zhang et~al.~\cite{zhang2010corner} presented an edge curve based corner detector where the derivative information between the position of the point on the edge curve and the position of the adjacent points are used to construct a gradient correlation matrix (GCM) for detecting corners. Teng et~al.~\cite{teng2015effective} extended the CPDA method~\cite{awrangjeb2008robust} and the ends of a given chord and the midpoint on the curve segment between the ends of the chord can be used to form a triangle. The ratio of the chord length to the sum of the other two arm lengths of the triangle is computed for detecting corners on edge contours. Ravindran and Mittal~\cite{ravindran2016comal} extended the maximally-stable extremal regions detector~\cite{Matas2004robust} to extract specific level lines. For each level line, the pixels on the line should have the same intensity. And then a matrix similar with GCM~\cite{zhang2010corner} is constructed for detecting corners on level lines. Mustafa et~al.~\cite{mustafa2018msfd} applied the multi-scale bilateral filters to segment an input image and then corners are extracted at the intersection of the borders of three or more regions.

\subsubsection{Intensity variation of edge pixels based interest point detection}
Interest point extraction algorithms based on intensity variations on edge pixels determine interest points by analyzing the intensity variation on edge pixels. Kitchen and Rosenfeld~\cite{kitchen1982gray} used the first and second order intensity derivative of an edge pixel $(x,y)$ as a corner measure $\Im(x,y)$
\begin{equation}\begin{aligned}
\label{eq13}
\jmath(x,y)=&I_{x x}(x,y)I_{y}(x,y)^2+I_{y y}(x,y)I_{x}(x,y)^2\\
        &-2I_{x y}(x,y)I_{x}(x,y)I_{y}(x,y),\\
\Im(x,y)=&\frac{\jmath(x,y)}{I_{x}(x,y)^2I_{y}(x,y)^2}.
\end{aligned}\end{equation}

Deriche and Giraudon~\cite{deriche1993computational} used a Gaussian filter to study the characteristics of a wedge and Y-type corners in a scale space, and corners are extracted on edge contours based on the corner properties. Parida et~al.~\cite{parida1998junctions} built a general junction model which can be used to represent an L-, Y-, X-, or star-type corners. And then the minimum description length technique~\cite{rissanen1983universal} is used to obtain the distribution of intensities at pixels within the region of the junction model for detecting and classifying junctions on edge contours. Fayolle et~al.~\cite{fayolle2000robustness} applied multi-scale wavelet to smooth the input image, and the points with the gradient direction changing rapidly on edge contours are marked as corners in the scale space. Sinzinger~\cite{sinzinger2008model} analyzed the properties of edge pixels based junction model and then the differences between the gray values of the pixels in a given region and the gray value of the center pixel of the region are added as a corner measure to detect corners on edge contours. In order to localize junctions and specify their corresponding orientations, Elias and Lagani{\`e}re~\cite{elias2011judoca} used the consistency of the gradient direction on edge pixels for detecting corners. Azzopardi and Petkove~\cite{azzopardi2012trainable} utilized the weighted geometric mean of the blurred and shifted responses of the selected Gabor filters for detecting keypoints on contours. To improve the affine robustness of the algorithm, Shui and Zhang~\cite{shui2013corner} employed the anisotropic Gaussian directional derivative (ANDD) filters~\cite{Shui2012} to derive the ANDD representations of the L-, Y-, X-, and star-type corners. Then the differences of the geometric shapes of the directional derivative between edges and corners are used for extracting corners on edge contours. Li et~al.~\cite{li2014scale} applied the change of the gradient direction as a corner measure for detecting corners on edge contours. To improve detection accuracy, Xia et~al.~\cite{xia2014accurate} applied a
contrario method~\cite{desolneux2000meaningful} based on the statistical model of gradient information in the neighborhood of feature points to replace the method about constructing linear scale space to obtain the scale information of junctions. To identify the discrepancies between corners and edges on edge contours, Zhang et~al.~\cite{zhang2014corner} used the magnitude responses of the imaginary part of the Gabor directional derivative filters to detect corners. In order to improve the noise robustness of the algorithm, Zhang and Shui~\cite{zhang2015contour} applied the distribution differences of gradient directions at edge pixels and corners for extracting corners on edge contours. In order to obtain stronger affine robustness and estimate the geometric structure of a junction, Xue et~al.~\cite{xue2017anisotropic} extended the ACJ method~\cite{xia2014accurate} by applying junction branch length estimation and constructing an anisotropic nonlinear scale space.

\subsection{Machine learning based IFI extraction techniques for interest point detection}
The existing machine learning based IFI extraction techniques for interest point detection include: first-order derivative based, first- and second-order derivative combination based, image pixel characteristic analysis based, and high level feature representation based methods.

Rosten et~al.~\cite{rosten2006machine, rosten2008faster} followed the SUSAN method~\cite{smith1997susan} and proposed the features from the accelerated segment test (FAST) algorithm based on the comparison of pixel values between a target point and a circular neighborhood template. Meanwhile, the decision tree technique~\cite{Quinlan1986Induction} was applied for speeding up corner detection. Lepetit and Fua~\cite{lepetit2006keypoint} extended the FAST method where the LoG filters~\cite{lindeberg1993detecting} were used to extract candidate interest points in the scale space, and then the randomized trees technique~\cite{Quinlan1986Induction} was used to select interest points from the candidates. Trujillo and Olague~\cite{trujillo2008automated} applied the random combination of forty-six basic operators (e.g., first- and second-order horizontal and vertical directional derivatives, autocorrelation matrix, and curvature) in the framework of genetic programming technique~\cite{banzhaf1998genetic} for training an interest point detector. The detector was required to have a high detection repeatability on interest points under image affine transformations. Verdie et~al.~\cite{verdie2015tilde} utilized the first-order horizontal and vertical directional derivatives of images under drastic imaging condition changes of weather and lighting for training a piece-wise linear regressor and performing interest point detection. Barroso-Laguna et~al.~\cite{Barroso-Laguna2019ICCV} followed the ideas of Harris~\cite{harris1988combined} and DoG methods~\cite{lowe2004distinctive}. The first- and second-order horizontal and vertical directional derivatives are used to train a detector according to the criterion that the detected interest points under image affine transformations should have a high detection repeatability. Hong-Phuoc et~al.~\cite{hong2018sck} applied a sparse coding technique to analyze the complexity of interest points on images, and an image pixel where the information complexity reaches a certain level is determined as an interest point.

Although it is indicated in~\cite{fong2017interpretable} that one cannot accurately determine how the black box operation extracts the required IFI, machine learning techniques are believed to have good capability to learn high-level image abstractions, have also been applied for interest point detection, and demonstrate appealing successes in interest point detection and interest points based computer vision tasks (e.g., interest points based image matching). In the following, we will illustrate the high-level semantic abstraction information based interest point detection methods in terms of supervised~\cite{richardson2013learning, yi2016lift, zhang2017learning, huo2020improved, noh2017large, di2018kcnn, detone2018superpoint, bhowmik2019reinforced, sarlin2019coarse,benbihi2019elf} and unsupervised~\cite{lenc2016learning, savinov2017quad, zhang2018learning, ono2018lf, shen2019rf, dusmanu2019d2, revaud2019r2d2, truong2019glampoints} based approaches.

The goal of supervised learning is to utilize input-label pairs to learn a function that predicts a label (or a distribution over labels) given the input~\cite{goodfellow2016deep}. Yi et~al.~\cite{yi2016lift} proposed a learned invariant feature transform (LIFT) method by training a convolutional neural network on image patches corresponding to the same feature but viewed under different ambient conditions. Zhang et~al.~\cite{zhang2017learning} extended the Lenc and Vedaldi method~\cite{lenc2016learning} by using features detected by the TILDE method~\cite{verdie2015tilde} as a guidance. Huo et~al.~\cite{huo2020improved} combined the score map extracted by a handcrafted detector (such as DoG) with the feature selection technique proposed in ~\cite{zhang2017learning} to improve the performance of a covariant local feature detector. Noh et~al.~\cite{noh2017large} proposed a detector and descriptor framework based on an attention network to detect key points and to describe them simultaneously. The framework first obtains the saliency response map of the image through the saliency convolution layer trained by feature matching based image retrieval. Detone et~al.~\cite{detone2018superpoint} introduced a self-supervised learning detector for extracting features from images. Bhowmik et~al.~\cite{bhowmik2019reinforced} extended the Superpoint method~\cite{detone2018superpoint}, and a reinforcement learning technique ~\cite{williams1992simple} is applied to a matching task based on local features (extracted by Superpoint ~\cite{detone2018superpoint}) for detecting interest points from an input image. Sarlin et~al.~\cite{sarlin2019coarse} applied the Mobile-Net~\cite{sandler2018mobilenetv2} and the decoder part of the Superpoint~\cite{detone2018superpoint} method for obtaining a final score map from an input image. The points with local extreme score values are determined as interest points. Benbihi et~al.~\cite{benbihi2019elf} applied a simple method in which the gradient responses of deep learning features are used for detecting interest points. Xu et al.~\cite{xu2020improved} applied dilated convolution to obtain a feature response map with multi-resolution for improving the scale invariant property of the detection method and solve the problem of detecting occluded interest points by adding tags of occluded interest points in the training set.

On the contrary, unsupervised learning methods do not require such labels or other targets~\cite{goodfellow2016deep}. To improve the robustness of interest point detection with image affine transformations, Lenc and Vedaldi~\cite{lenc2016learning} proposed a covariant constraint to transform interest point detection to a regression processing, which applied a trained local transformation predictor for replacing a learned score map to indicate the probability of local features. The Quad-networks~\cite{savinov2017quad} method is the first unsupervised interest point detection method, not relying on a hand-crafted feature detector for the training set generation. The training process is based on the rank robustness of the corresponding point sets under image affine transformations. A point with local extrema response score is defined as an interest point. Zhang et~al.~\cite{zhang2018learning} extended the method in~\cite{savinov2017quad} for detecting interest points in texture images. Ono et~al.~\cite{ono2018lf} trained a neural network using the high repeatability of corresponding point pairs to obtain multi-scale feature score maps, and interest points are determined as the points with local maximum scores. Shen et~al.~\cite{shen2019rf} proposed an end-to-end trainable matching network based on a receptive field, and sparse correspondences between images are computed for presenting a new interest point detector. Dusmanu et~al.~\cite{dusmanu2019d2} proposed a framework to accomplish feature detection and description by looking for the local maxima in the response score map of the fourth convolutional layer of VGG-16 pre-trained on ImageNet~\cite{deng2009imagenet}. To improve the repeatability and reliable performance for image matching, Revaud et~al.~\cite{revaud2019r2d2} proposed a learning based model trained by maximizing the cosine similarity between corresponding image patchs for detecting interest points. Luo et~al.~\cite{luo2020aslfeat} applied the light-weight L2-Net~\cite{tian2017l2} to replace the VGG backbone~\cite{simonyan2014very} used in D2-Net~\cite{dusmanu2019d2} and deformable convolutional networks~\cite{zhu2019deformable} to obtain the score map of interest points. Meanwhile, the non-maximum suppression technique was used to obtain the final feature point sets. Yan et~al.~\cite{yan2021unsupervised} proposed equivalent probability formula of the properties (sparsity, repeatability, invariability, discriminability, and information complexity) about the interest point and its corresponding neighborhood region to build a model loss function. Meanwhile, in order to fit the above five non-differentiable properties, the expectation maximization and efficient approximation techniques were applied to replace the commonly used gradient descent based model optimization methods. Finally, the combination between the above five non-differentiable properties and the Superpoint~\cite{detone2018superpoint} model were used to obtain an interest point detector and a descriptor.

\subsection{Frequency domain based IFI extraction techniques for interest point detection}
Frequency domain based interest point detection methods extract frequency information for detecting interest points. Kovesi~\cite{kovesi2003phase} applied the phase congruency information to construct a second-order moment matrix in the frequency domain for detecting corners. Chen et~al.~\cite{chen2014low} proposed a block-wise scale-space representation method to decrease the computation complexity of the LoG~\cite{lindeberg1993detecting} and DoG~\cite{lowe2004distinctive} methods. An input image is decomposed into blocks, and then the frequency domain convolution and parallel implementation are used for extracting interest points in each block of the images.

\section{Datasets and performance evaluation}
In this section, the popular datasets for interest point detection are firstly presented. Secondly, the existing performance evaluation criteria for interest point detection are introduced. Thirdly, the detection performances for the most representative methods are evaluated, discussed, and summarized.

\subsection{Image datasets}
Datasets have played a very important role for interest point detection. Datasets not only are a means of evaluating the performances of interest point detection algorithms, but also help advance the field of interest point detection greatly.

The attributes of popular datasets for performance evaluation on interest point detection are summarized in Table~\ref{stest1}. Moreover, the download links of these datasets~\cite{mikolajczyk2005comparison, rosten2006machine, strecha2008benchmarking, aanaes2012interesting, zitnick2011edge, wong2011dynamic, hauagge2012image, heinly2012comparative, shui2013corner, verdie2015tilde, moo2016learning, balntas2017hpatches} can be found in~\cite{vgg, FAST, LDAHash, DTUimage, EF, AdelaideRMF, Symbench,  Heinly, Webcam,Viewpoint,HPatches}.

Currently, the most popular evaluation criterion on interest point detection is whether an interest point detector can accurately detect corresponding interest points in the image pairs under image affine transformations. The input for the evaluation system only needs the position and scale information for the detected interest points from an input image. Without the influence of a descriptor, an accurate evaluation result about comparing the localization error in a corresponding interest point pair can be obtained by the evaluation criterion of image affine transformations. It is worth to note that it is still a tedious and complicated problem for verifying the accuracy of the matching between interest point pairs. A general practice is to artificially obtain the ground truth about the transformation parameter matrix, the homography, which is used to describe the geometric correspondence between the original image and the transformed image with different viewpoints. The representatives of this evaluation system are proposed in~\cite{mikolajczyk2005comparison, schmid2000evaluation}. The VGG dataset~\cite{mikolajczyk2005comparison,vgg}, which includes transformed conditions with zoom, rotation, image blur, viewpoint change, light change, and JPEG compression, was proposed to provide data support for evaluation. However, the data volume of the original VGG image set~\cite{mikolajczyk2005comparison,vgg} is too small, which limits the fairness and objectivity on evaluation results. To solve this problem, many datasets~\cite{strecha2008benchmarking, zitnick2011edge, wong2011dynamic, hauagge2012image, heinly2012comparative, verdie2015tilde, moo2016learning, balntas2017hpatches} that focus on seasonal changes, light changes, automatic exposure, and image quality changes have been proposed to expand the dataset library for performance evaluation on interest point detection under different image transformations.

It is worth to note that the aforementioned datasets can also be applied well on some improved repeatability evaluation criteria. In~\cite{awrangjeb2012performance}, repeatability and localization error are used to determine the affine invariance and localization accuracy for interest point detectors. In~\cite{fraundorfer2005novel}, the methods proposed in~\cite{mikolajczyk2005comparison}  improved for non-planar scenes. The trifocal tensor and the point transfer property were applied to obtain the ground truths of 3D scenes. In~\cite{zitnick2011edge}, the entropy measure of a detector was utilized to reduce the impact of overlapping points on performance evaluation. In~\cite{rosten2006machine}, the method in~\cite{schmid2000evaluation} was improved by replacing the planar scene dataset with 3D scenes, which allows the evaluation criteria to have the ability to assess the features caused by geometry. The ground truth of a 3D model is manually aligned to the scene. Then a mixture of simulated annealing and gradient descent is applied to minimize the error about the sum of squared differences between all pairs of frames and reprojections. Meanwhile, corresponding datasets with viewpoint change, scale change, and radial distortion were provided. Aan{\ae}s et~al.~\cite{aanaes2012interesting} proposed a large-scale dataset by applying 3D structured light scaning, precision manipulator, and an environment without light interference, which can be applied in the corresponding evaluation criteria based on recall.

\renewcommand{\arraystretch}{1.5}
\begin{table*}[htbp]
\setlength{\abovecaptionskip}{5pt}
\setlength{\belowcaptionskip}{0pt}

\caption{Popular datasets for performance evaluation on interest point detection.}
\label{stest1}
\centering
\begin{tabular}{lllllll}
\hline
Dateset name& Total images& Categories &Image size& Started year&Highlights\\
\midrule[1pt]
VGG~\cite{mikolajczyk2005comparison}&48&8&1,000$\times$7,00&2005&Eight image scenes including `Bark',    \\
~&~&~ &800$\times$640&~&`Bikes', `Boat', `Graffiti', `Leuven', `Trees',   \\
~&~&~ &765$\times$512&~&`Ubc', and `Wall' with 6 images in each sence.  \\
~&~&~&921$\times$614&~& \\
Rosten~\cite{rosten2006machine}&37&3  &768$\times$576&2006&Three sequence, `Box', `Maze', and `Junk',     \\
~&~&~ &&~&with plenty of registered images in each sequence. \\
Strecha~\cite{strecha2008benchmarking}&19&2 &3,072 $\times $ 2,028&2008&Two scenes, Fountain-P11 and Herz-Jesu-P8,  \\
~&~&~&~&~&with a total of 19 three-dimensional images provided. \\
DTU~\cite{aanaes2012interesting}&135,660&60&1,200 $\times$ 1,600&2011&Sixty scenes with $119 \times 19$ images in each scene.  \\
EF~\cite{zitnick2011edge}&38&5&about $600\times 600$ &2011&Five sence including `Yosemite','Obama',    \\
~&~&~&&~&`Mt.Rushmore', `Notre Dame', and `Painted Ladies'.   \\
Ade-RMF~\cite{wong2011dynamic}&76&38&$640 \times 480$&2011&There are 38 scenes with 76 images shooting  \\
~&~&~&2,272 $\times$ 1,704&~&on viewpoint change provided.  \\
Symbench~\cite{hauagge2012image}&92&46 &$300 \times 400$ to&2012& There are 46 image scenes with light change \\
~&~&~& 1,000 $\times$ 1,000&~&in this dataset.\\
Heinly~\cite{heinly2012comparative}&65&7&1,024 $\times$ 768&2012& There are 7 scenes with 65 images shooting on pure \\
~&~&~&$600 \times 800$&~& rotation, pure scaling, illumination changes,     \\
~&~&~&~&~&white balance, auto-exposure, and image quality.\\
Webcam~\cite{verdie2015tilde}&120&6&$316 \times 240$ to&2015& There are six scenes, including `StLouis',  \\
~&~&~&2,048 $\times$ 1,536&~&`Mexico', `Chamonix', `Courbevoie', `Frankfurt',  \\
~&~&~&~&~&and ` Panorama', with 120 pictures in total.  \\
Viewpoints~\cite{moo2016learning}&30&5&1,000 $\times$ 700&2016&There are 5 scence with 30 images in total shooting   \\
~&~&~&700 $\times$ 1,000&~&on viewpoint changes and rotations. \\
HPatches~\cite{balntas2017hpatches}&1,856&116 &65 $\times$ 30,000 to&2017&There are 116 scenes with 16 image in each set shooting  \\
~&~&~&65 $\times $130,000&~& on viewpoint change and  photometric changes. \\
\hline
\end{tabular}
\end{table*}

\subsection{Evaluation criteria}
The existing evaluation criteria on interest point detection~\cite{Trajkovi1998Fast, schmid2000evaluation, mohanna2001performance, rosten2006machine, awrangjeb2012performance, aanaes2012interesting, dickscheid2011coding, mikolajczyk2005comparison, fraundorfer2005novel, philbin2007object, zitnick2011edge, gauglitz2011evaluation, jiang2013performance, aanaes2016large, mustafa2018msfd, fan2019performance} can be divided into four groups: detection capability and localization accuracy metrics for interest point detectors on test images with ground truths~\cite{mohanna2001performance, dickscheid2011coding}, performance evaluation on interest point detectors in the field of visual applications~\cite{Trajkovi1998Fast, gauglitz2011evaluation, fan2019performance}, repeatability metrics for interest point detectors under image affine transformations~\cite{schmid2000evaluation, rosten2006machine, awrangjeb2012performance, aanaes2012interesting, mikolajczyk2005comparison, fraundorfer2005novel, zitnick2011edge}, and execution time and memory usage~\cite{zhang2019corner2, zhang2020corner, truong2019glampoints, kong2013generalized, elias2011judoca, alcantarilla2012kaze, zhang2007multi, mustafa2018msfd, awrangjeb2012performance, mainali2011robust, bay2006surf, verdie2015tilde, mainali2013sifer}.

The detection capability and localization accuracy metrics are often used for performance evaluation on corner detection methods~\cite{zhu1995critical, zhang2020corner, zhang2019corner2, wang2020corner, su2012junction, bennett2014chess, zhang2009robust, shui2013corner, xia2014accurate, zhang2007multi, he2008corner, pedrosa2010anisotropic, zhang2019discrete, zhang2010corner, elias2011judoca,zhang2014corner, zhang2015contour}. The advantage of this evaluation criterion is that it allows researchers to intuitively determine whether real corners are detected and what the accuracy on corner localization is. However, it is indicated in~\cite{awrangjeb2008robust} that it is difficult to use ground truth based evaluation techniques for assessing the performances of detectors in a large database with thousands of test images. The reason is that it is impossible for users to mark reference corners one by one in thousands of test images.

Another method for evaluating the performances of interest point detection algorithms is to assess the performances on the interest point detectors in computer vision tasks such as target tracking~\cite{Trajkovi1998Fast, gauglitz2011evaluation} or image matching~\cite{agrawal2008censure}. It is worth to note that in addition to the performances of the interest point detection algorithms, the performances of these vision tasks are also greatly affected by the descriptors.

Considering that interest point detectors need to have strong repeatability and real time performances for the application of certain vision tasks (e.g., simultaneous localization and mapping) which means that they should have high probabilities to find the local correspondences between different images within a specified time. So, most performance evaluation criteria on interest point detectors are designed to evaluate the repeatability~\cite{schmid2000evaluation, rosten2006machine, awrangjeb2012performance, aanaes2012interesting, mikolajczyk2005comparison, fraundorfer2005novel, zitnick2011edge} and execution time~\cite{zhang2019corner2, zhang2020corner, truong2019glampoints, alcantarilla2012kaze, zhang2007multi, mustafa2018msfd, awrangjeb2012performance, mainali2011robust, bay2006surf, verdie2015tilde, mainali2013sifer}.

There are three popular repeatability criteria for evaluating the performances of interest point detection algorithms: recall rate under the DTU-Robots dataset~\cite{aanaes2012interesting},repeatability score under region repeatability evaluation~\cite{mikolajczyk2005comparison}, and average repeatability under image affine transformations~\cite{awrangjeb2008robust, awrangjeb2012performance}. In addition, the matching score~\cite{mikolajczyk2005comparison} based on the detected interest point pairs and the distances between the corresponding descriptors is applied to evaluate the matching performance of the detector. These four criteria~\cite{aanaes2012interesting, mikolajczyk2005comparison, awrangjeb2012performance,mikolajczyk2005comparison} are briefly introduced in the following.

\subsubsection{Repeatability metric under the DTU-Robots dataset}

The recall rate on the DTU-Robots dataset~\cite{aanaes2012interesting} is utilized to assess the performances of interest point detection algorithms. The dataset contains 60 scenes and 135,660 color images with 1,200$\times$1,600 pixels which are obtained with changes in viewpoint, scale, and lighting condition. For each scene, a camera is placed at 119 locations in three transverse arc routes (Arc 1, Arc 2, and Arc 3) and a straight route (Linear path) in front of the scene for acquiring different images. The central frame nearest to the scene is selected as the key frame image. In the first stage, interest points extracted from each test image and the key frame image are compared. In addition, the scene is illuminated by 19 independently controlled LEDs for imitating the natural scenes. These 19 LEDs can be put together to simulate different light changes from the diffuse reflection of a cloudy day to the highly directional condition with sunlight. In the second stage, the scene relighting has been performed both from right to left and from back to front to assess the robustness of interest point detection algorithms with changes of lightings. For each camera location, ten different lighting settings can be configured by altering the lighting direction and obtain ten different corresponding images. Feature points obtained from each test image and the reference image are compared (the tenth image is selected as the reference image in this stage).

For a pair of images, their corresponding recall rate is used as a performance measure
\begin{equation}\begin{aligned}
\label{eq56}
{{\tau_{\text{metric}}}} = \frac{{\zeta_{\text{corresp}}}}{{\zeta_{\text{total}}}},
\end{aligned}\end{equation}
where $\zeta_{\text{corresp}}$ is the number of matching points between the key frame image and the image at each different location, and $\zeta_{\text{total}}$ is the number of detected interest points in the key frame image. If an interest point in the key frame image meets the following three conditions, it is selected as a matching point.

$\bullet$ Epipolar Geometry: Precise setting of each camera position provides the foundation for the relationship between interest points obtained from images with known epipolar geometry. If one interest point is more than 2.5 pixels away from the corresponding epipolar line, the interest point will be eliminated.

$\bullet$ Surface Geometry: If the 3D position of the two matched interest points is within a 3D cube with a side length of 10 pixels from the scene surface obtained from the structured light reconstruction, it is regarded as a positive matching. Instead, the interest points with a 3D reconstruction which is outside the cube with a diameter of 10 pixels are discarded.

$\bullet$ Absolute Scale: The corresponding region size of a detected interest point from the key frame image and the corresponding region size of the associated detected interest point in another test image must be within a ratio of 2.

\subsubsection{Repeatability score under region repeatability evaluation}
In~\cite{mikolajczyk2005comparison}, each of the image sequences used in the evaluation contain 6 images of naturally textured scenes with increasing geometric and photometric transformations. The images in a sequence are related by a homography which is provided with the image data. The repeatability for a given pair of images is computed as the ratio between the number of region-to-region correspondences and the minimum number of regions in one of the images. Two regions are deemed to correspond if the overlap error $\varsigma$ is sufficiently small. In this evaluation criteria~\cite{mikolajczyk2005comparison}, the overlap error is defined as 1 minus the ratio between the intersection of regions $A\cap H^{\top}BH$ and the union of regions $A\cup H^{\top}BH$
\begin{equation}\begin{aligned}
\varsigma=1-\frac{A\cap H^{\top}BH}{A\cup H^{\top}BH},
\label{eq55}
\end{aligned}\end{equation}
where $A$ represents a region in the original image, $B$ represents another region in the transformed image, and $H$ is the homography between the original image and the transformed image. When the overlap error between two regions is less than 0.4, a correspondence is detected. The repeatability score for a given pair of images is computed as
\begin{equation}\begin{aligned}
\label{eq56}
RS_i = \frac{CR_{1i}}{\text{min}(C_1,C_i)},
\end{aligned}\end{equation}
where $CR_{1i}$ is the number of correspondences between the original image and the $\emph{i}$-th transformed image ($i=1,\mathellipsis,6$), $C_1$ is the number of the detected interest points from the original image, and $C_i$ is the number of the detected interest points from the $\emph{i}$-th transformed image.

\subsubsection{Repeatability under image affine transformation}
In~\cite{awrangjeb2008robust}, the average repeatability $\xi_{\text{avg}}$ and localization error $\vartheta_{\text{error}}$ explicitly measure the performance of a detector by comparing the detected interest points from the original image and the detected interest points from the affine transformed image. The definition of the average repeatability is
\begin{equation}
{
\xi_{\text{avg}}=\frac{P_{Nm}}{2}\left(\frac{1}{P_{No}}+\frac{1}{P_{Nt}}\right),
}
\end{equation}
where $P_{No}$ is the number of extracted interest points from the original image, $P_{Nt}$ is the number of extracted interest points from the transformed image, and $P_{Nm}$ is the number of matched interest points between the original and the transformed images. For an interest point $\mu$$=$$(x_{o},y_{o})$ extracted from the original image, its corresponding location in the transformed image is point $\nu$$=$$(x_{t},y_{t})$. If one interest point is extracted from the transformed image and its location is near point $\nu$$=$$(x_{t},y_{t})$ (e.g., within 4 pixels), it means that a pair of matched interest points is obtained.

The localization error is defined as the average distance of all matched interest point pairs between the original and the transformed images. Let $\{(\hat{x}_{p},\hat{y}_{p}),(x_p,y_p)$$:$$~p=1,2,\mathellipsis,P_{Nm}\}$ be the matched interest point pairs. Then the localization error $\vartheta_{\text{error}}$ is
\begin{equation}
{
\vartheta_{\text{error}}=\sqrt{\frac{1}{P_{Nm}}\sum_{p=1}^{P_{Nm}}((\hat{x}_{p}-x_p)^{2}+(\hat{y}_{p}-y_p)^{2})}.
\label{AL}
}
\end{equation}

\subsubsection{Matching score under the VLBenchmarks}
In~\cite{mikolajczyk2005comparison}, Mikolajczyk et al. presented a criteria based on image matching score which is denoted in VLBenchmarks~\cite{lenc12vlbenchmarks}. The matching score is calculated in two steps.

Firstly, two regions are deemed to match if the overlap error $\varepsilon$ is sufficiently small. For region repeatability evaluation~\cite{mikolajczyk2005comparison}, the overlap error is defined as one minus the ratio between the intersection of regions, $A\cap H^{\top}BH$, and the union of the regions, $A\cup H^{\top}BH$,
\begin{equation}\label{eq15}
\varepsilon=1-\frac{A \cap H^{T} B H}{A \cup H^{T} B H},
\end{equation}
where $A$ represents a region in the original image, $B$ represents the corresponding region in the transformed image, and $H$ is the corresponding homography between the original and the transformed images. When the overlap error between the two regions is less than $40\%$, a correspondence is detected.

Secondly, the matching score is calculated as
\begin{equation}\label{eq14}
MS_{\mathrm{i}}=\frac{MP_{i}}{\min \left(NP_{1}, NP_{i}\right)},
\end{equation}
where $MP_{i}$ is the number of real matched interest point pairs between the original image and the $i$-th transformed image ($i=1, ... ,6$). And $NP_{1}$ is the number of interest point detected in the original image. Meanwhile, $NP_{i}$ is the interest point number detected from the $i$-th transformed image. A match is defined as the nearest neighbour in the descriptor space. The descriptors are compared with a given distance measure (e.g., the Euclidean distance).

\subsection{Experimental Results}
Comprehensive performance evaluation of eighteen state-of-the-art methods (FAST~\cite{rosten2006machine}, Harris~\cite{harris1988combined}, ACJ~\cite{xia2014accurate}, Harris-Laplace~\cite{mikolajczyk2004scale},  DoG~\cite{lowe2004distinctive}, SURF~\cite{bay2006surf}, KAZE~\cite{alcantarilla2012kaze}, FOAGDD~\cite{zhang2020corner}, SOGGDD~\cite{zhang2019corner2}, CPDA~\cite{awrangjeb2008robust}, ANDD~\cite{shui2013corner}, GCM~\cite{zhang2010corner}, New-Curvature~\cite{zhang2019discrete}, LIFT~\cite{yi2016lift}, D2-Net~\cite{dusmanu2019d2}, LF-Net~\cite{ono2018lf}, He $\&$ Yung~\cite{he2008corner}, and DT-CovDet~\cite{zhang2017learning}) are reported in this section. In addition, four popular algorithms on descriptor including BRIEF~\cite{calonder2010brief}, SIFT~\cite{lowe2004distinctive}, Hard-Net++~\cite{HardNet2017}, and L2-Net~\cite{tian2017l2} are used in image matching evaluation. The DTU-Robots dataset~\cite{aanaes2012interesting} that contains 3D objects under changing viewpoints for recall rate evaluation is utilized to measure the average repeatabilities of the eighteen methods. One thousand images~\cite{deng2009imagenet} which contain different types of scenes without ground truths are utilized to measure the average repeatabilities of the eighteen methods under different image affine transformations, image JPEG compressions, and image noise degradations. Furthermore, the HPatches dataset~\cite{balntas2017hpatches} (image sets not patch sets) is applied on image matching evaluation. Furthermore, execution times are also investigated for the eighteen methods. The original codes for twelve of these methods in~\cite{shui2013corner, xia2014accurate, yi2016lift,  dusmanu2019d2, ono2018lf, zhang2020corner, zhang2019corner2, awrangjeb2008robust,zhang2017learning, zhang2010corner, zhang2019discrete, he2008corner, zhang2017learning} and two descriptor algorithms (Hard-Net++~\cite{HardNet2017} and L2-Net~\cite{tian2017l2}) are from the authors. The codes for the Harris-Laplace~\cite{mikolajczyk2004scale}, DoG~\cite{lowe2004distinctive}, and SIFT decriptor methods are from~\cite{vedaldi08vlfeat}. The code for the Harris method~\cite{harris1988combined} is from~\cite{cornerharris}. The codes for the FAST~\cite{rosten2008faster}, SURF~\cite{bay2006surf}, and KAZE~\cite{alcantarilla2012kaze} methods are from MATLAB. The codes for the BRIEF~\cite{calonder2010brief} algorithm is from Python-OpenCV.

\subsubsection{Repeatability metric on the DTU-Robots dataset}
In the DTU-Robots dataset~\cite{aanaes2012interesting}, fifty-four sets of images (a total of 122,094 test images) from the original sixty sets of images~\cite{aanaes2012interesting} are obtained for our evaluation (the 31th-36th sets cannot be downloaded from~\cite{aanaes2012interesting}). In this experiment, the threshold for each detector is adjusted so that each detector extracts about 1,000 interest points from each input image. The average match percentage for 119 positions are shown in the first row of Fig.~\ref{fig14}. The average repeatability metrics for the changes in light directions from right to left for four camera positions (1, 20, 64, and 65) are shown in the second row of Fig.~\ref{fig14}. The average repeatability metrics for the changes in light direction from back to front for four camera positions (1, 20, 64, and 65) are shown in the third row of Fig.~\ref{fig14}. The average repeatability metric for each detector is summarized in Table~\ref{test}. It can be found from Table~\ref{test} that the FOAGDD method~\cite{zhang2020corner} and the SOGGDD method~\cite{zhang2019corner2} achieved the best and the second best detection performance in this evaluation criteria respectively.
\renewcommand{\arraystretch}{1.3}
\begin{table}[!t]
\setlength{\abovecaptionskip}{0pt}
\setlength{\belowcaptionskip}{0pt}
\footnotesize

\caption{Average match percentage.}
\label{test}
\centering
\setlength{\tabcolsep}{4mm}{
\begin{tabular}{cccc}

\toprule
\multirow{3}{*}{Detector} &\multirow{2}{*}{119}    & \multicolumn{2}{c}{Light changes}       \\
\cline{3-4}
~~& \multirow{2}{*}{positions}  & Right  &  Back   \\
~~&   &  to left  & to front  \\
\midrule
 Harris & 0.724 & 0.614 & 0.549 \\

Harris-Laplace & 0.795 & 0.577 & 0.522\\

FOAGDD & \textbf{0.872} & \textbf{0.708} & 0.554 \\
\hline

SOGGDD& 0.842 & 0.654 & \textbf{0.704} \\

DoG& 0.728 & 0.563 & 0.486\\

SURF & 0.788 & 0.636 & 0.544\\

KAZE & 0.814 & 0.633 & 0.561\\
\hline
FAST &0.841 & 0.659& 0.611  \\

\hline
GCM & 0.716 & 0.550 & 0.529\\

CPDA & 0.713 & 0.559 & 0.535\\
ANDD & 0.633& 0.467 & 0.466\\

ACJ & 0.709 & 0.597 & 0.501\\

New-Curvature & 0.642 & 0.486 & 0.471\\

He \& Yung & 0.643 & 0.458 & 0.353\\
\hline
D2-Net & 0.774 & 0.651 & 0.677 \\

LF-Net & 0.797 & 0.663 & 0.681\\

LIFT & 0.760 & 0.578 & 0.633\\

DT-CovDet & 0.667 & 0.528 & 0.569\\

\bottomrule

\end{tabular}}
\end{table}

\begin{figure*}[!htbp]
\setlength{\abovecaptionskip}{0pt}
\setlength{\belowcaptionskip}{0pt}
\centering
\includegraphics[width=6.8in]{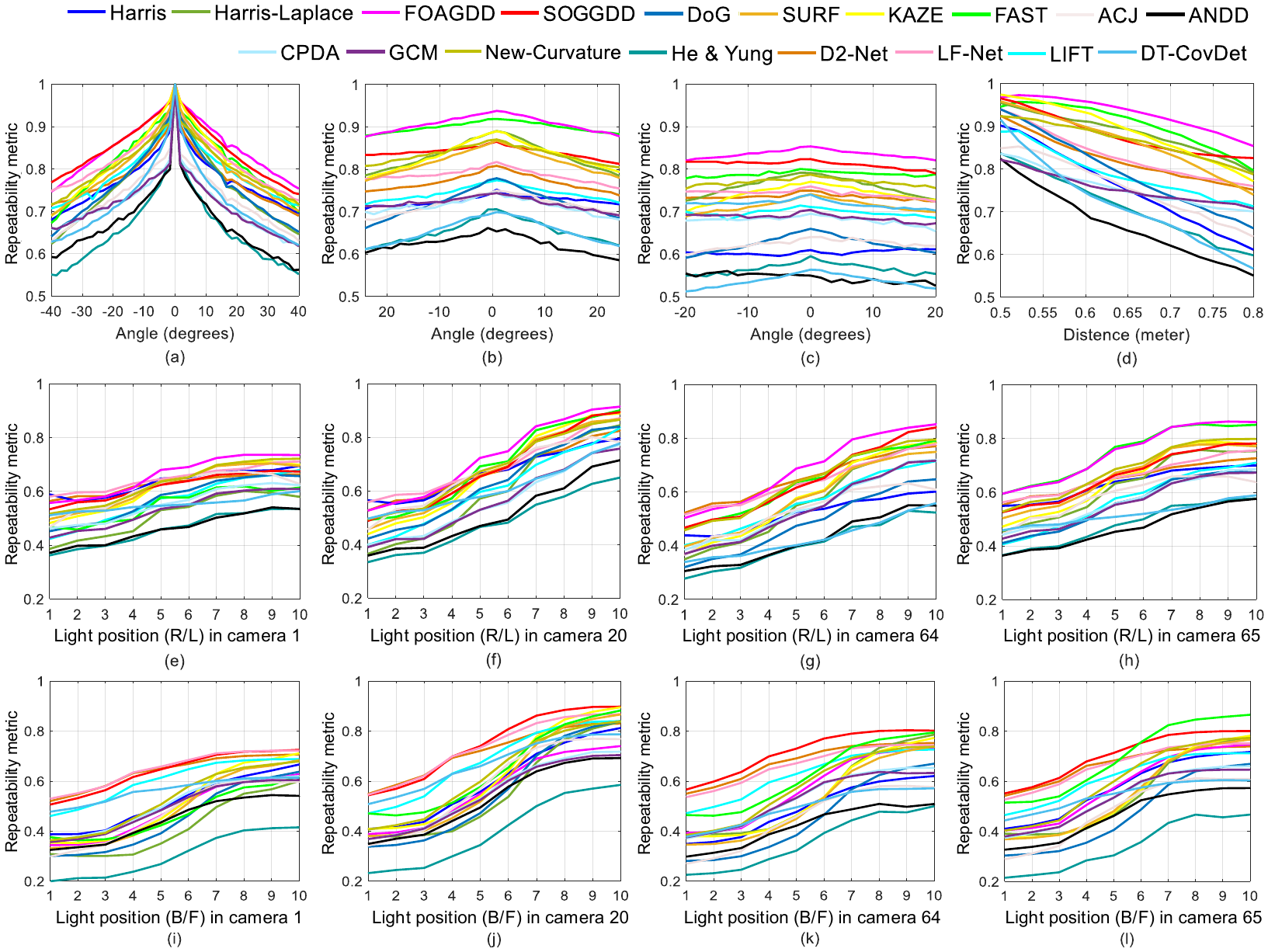}
\caption{Average repeatability metric for different scene settings with 119 positions and the changes in light directions for four camera positions (1, 20, 64, and 65). The first row is the results on position evaluation (a-d) Arc 1, Arc 2, Arc 3, and Linear path. The second row (e-h) is the average repeatability metrics for the changes in light directions from right to left (R/L). The third row (i-l) is the average repeatability metrics for the changes in light directions from back to front (B/F).}
\label{fig14}
\end{figure*}

\subsubsection{Repeatability score on region repeatability evaluation}
\begin{figure*}[!htbp]
\setlength{\abovecaptionskip}{5pt}
\setlength{\belowcaptionskip}{0pt}
\centering
\includegraphics[width=6.8in]{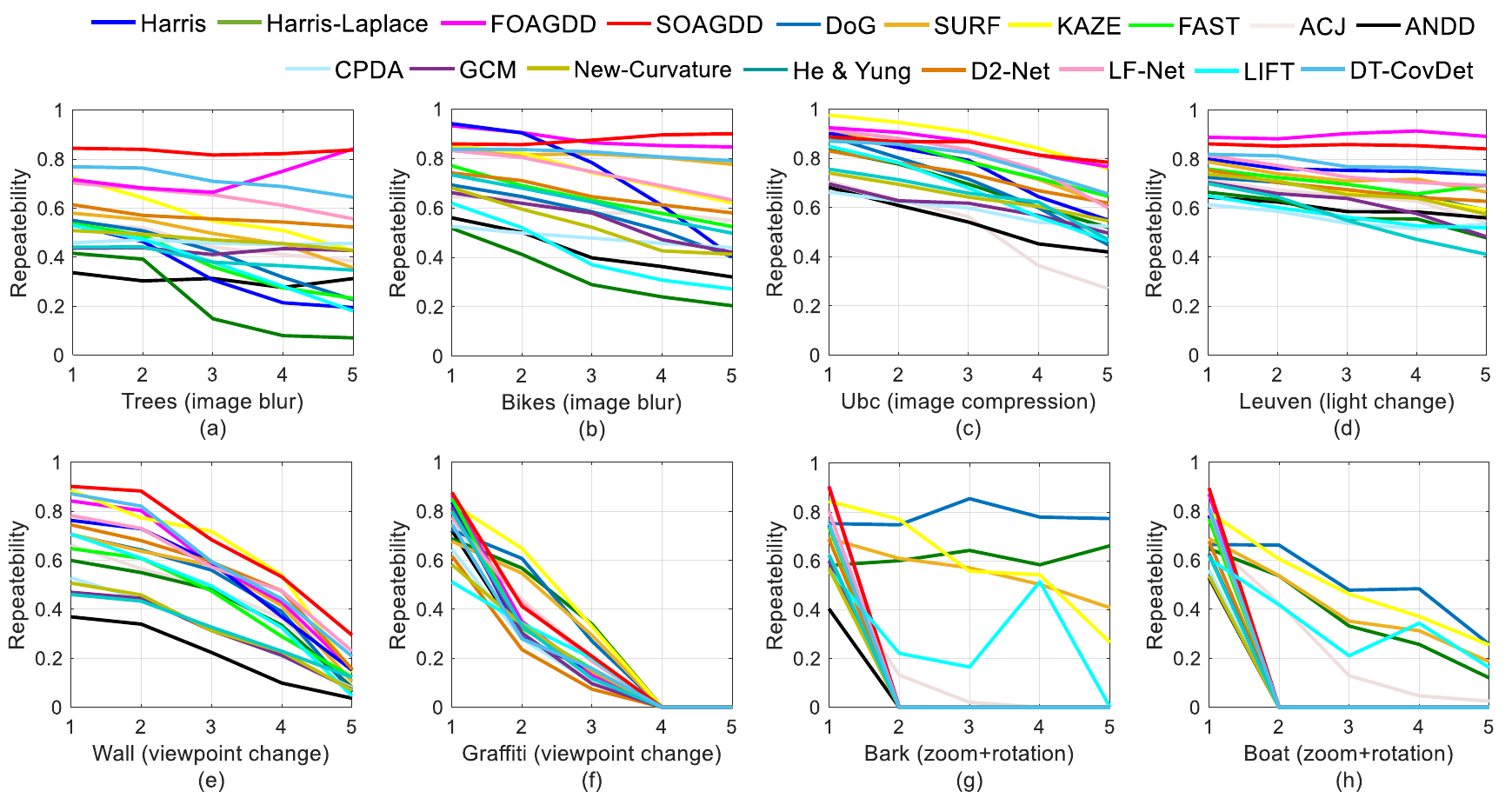}
\caption{Average repeatability of different interest point detectors on eight image scenes.}
\label{fig11}
\end{figure*}

In~\cite{mikolajczyk2005comparison}, an overlap based evaluation metric and eight image scenes (`Bark', `Bikes', `Boat', `Graffiti', `Leuven', `Trees', `Ubc', and `Wall') in the Oxford image database were proposed. The threshold for each method is tuned to extract about 1,000 interest points from each input image. The repeatability scores for the eight image sequences are illustrated in Fig.~\ref{fig11}.

\begin{figure*}[!htbp]
\setlength{\abovecaptionskip}{0pt}
\setlength{\belowcaptionskip}{0pt}
\centering
\includegraphics[width=6in]{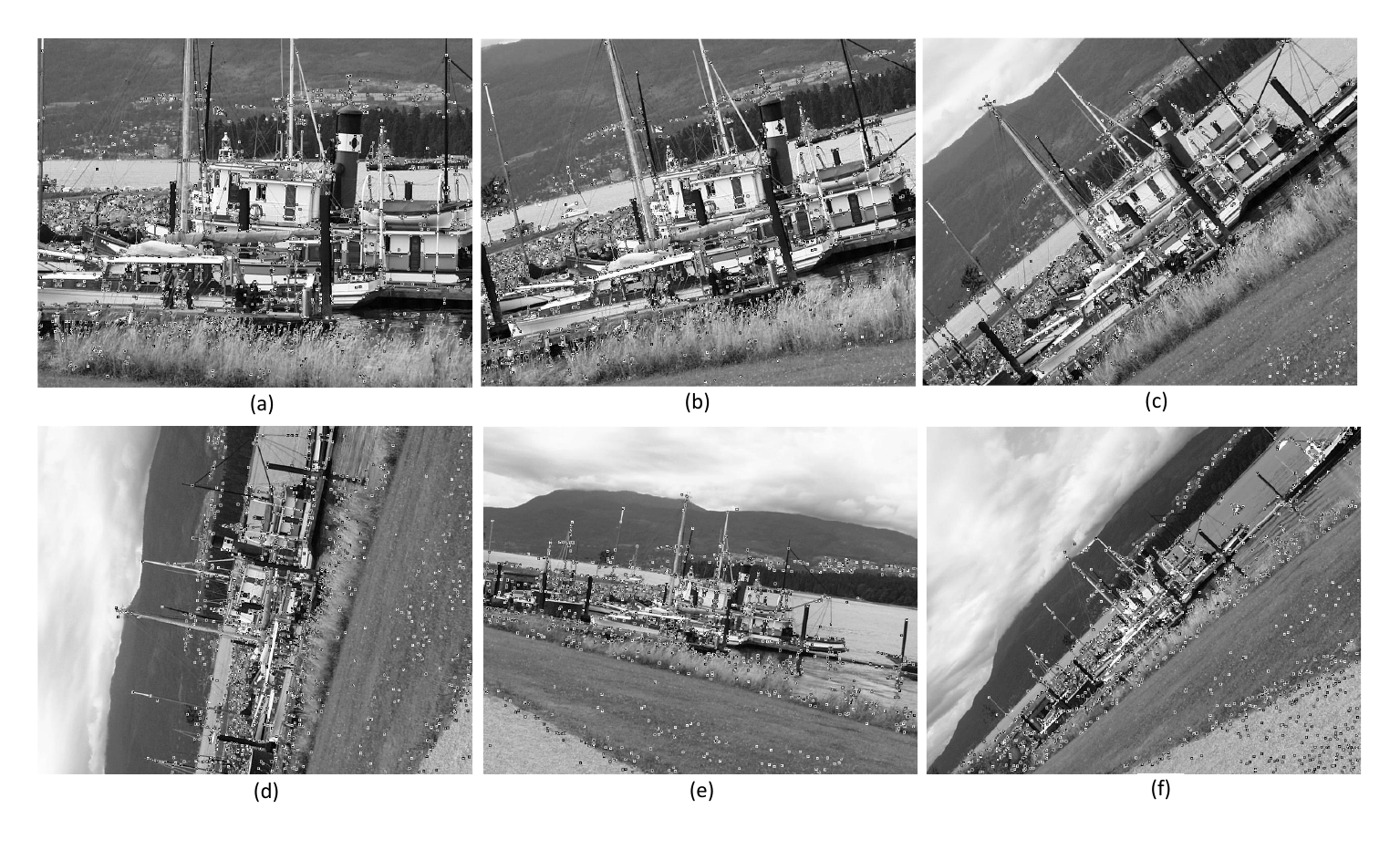}
\caption{The interest point detection results from the FOAGDD method on the `Boat' image sequence.}
\label{figboat}
\end{figure*}

\begin{figure*}[!htbp]
\setlength{\abovecaptionskip}{0pt}
\setlength{\belowcaptionskip}{0pt}
\centering
\includegraphics[width=6in]{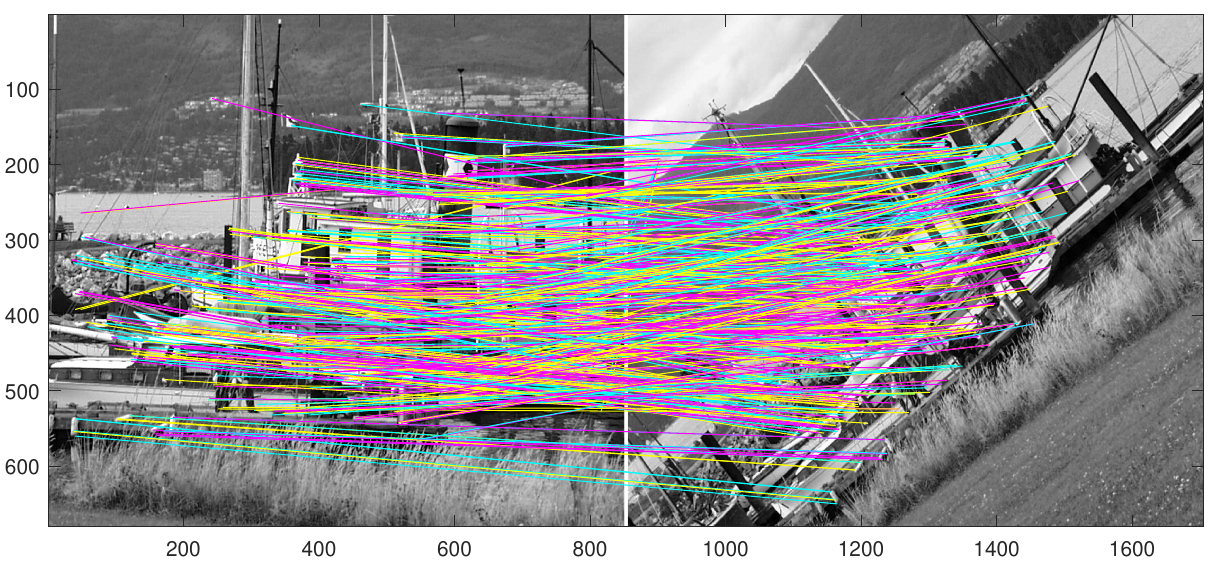}
\caption{The image matching result on the first image and the third image of the `Boat' image sequence using the FOAGDD method.}
\label{figboatmatching}
\end{figure*}

It is worth to note that the performance of single-scale methods (Harris~\cite{harris1988combined}, FAST~\cite{rosten2008faster}, ANDD~\cite{shui2013corner}, FOAGDD~\cite{zhang2020corner}, SOGGDD~\cite{zhang2019corner2}, CPDA~\cite{awrangjeb2008robust}, GCM~\cite{zhang2010corner}, New-curvature~\cite{zhang2019discrete}, and He $\&$ Yung~\cite{he2008corner}) cannot be tested with the `zoom+rotation' criterion in this performance evaluation. Fig.~\ref{fig11}(g) and (h) show the repeatability scores for the two image sequences on `Bark' and `Boat'. It can be seen that the number of correspondences between the reference image and the third image to the sixth image is zero. The reason is that the Oxford dataset is used to evaluate the performance of the affine region detectors. This evaluation mechanism in the `zoom + rotation' criterion is not applicable to the single-scale interest point detectors. Take the FOAGDD method~\cite{zhang2020corner} as an example, the interest point detection results of the FOAGDD method on the `Boat' image sequence are shown in Fig.~\ref{figboat}. It is worth to note that the threshold for the FOAGDD method is adjusted so that the detector extracts about 1,000 interest points from each input image. The image matching result for the first image and the third image of the FOAGDD method is shown in Fig.~\ref{figboatmatching}. A total of 601 pairs of matching are obtained using the detected corners. However, the number of correspondences between the first image and third image is zero under this evaluation criteria~\cite{mikolajczyk2005comparison}. The reason is that this evaluation mechanism for the `zoom + rotation' criterion~\cite{duval2015edges} usually needs an appropriate descriptor to handle large image zooming and rotations.

Based on the aforementioned analysis, six image sequences from~\cite{mikolajczyk2005comparison} are used for our evaluation and two image sequences (large zooming and rotations) are discarded in this experiment. The average repeatability score for each detector is summarized in Table~\ref{test3}. It can be found from Table~\ref{test3} that the FOAGDD method~\cite{zhang2020corner}, the SOGGDD method~\cite{zhang2019corner2}, and the KAZE method~\cite{alcantarilla2012kaze} achieved two optimal detection performance indicators in this evaluation criteria respectively.

\renewcommand{\arraystretch}{1.3}
\begin{table}[!htbp]
\setlength{\abovecaptionskip}{0pt}
\setlength{\belowcaptionskip}{0pt}
\footnotesize

\caption{Average result for repeatability under overlap rate.}
\label{test3}
\centering
\setlength{\tabcolsep}{1.7mm}{
\begin{tabular}{ccccccc}

\toprule
\multirow{2}{*}{Detector} & \multicolumn{6}{c}{Image}\\
\cline{2-7}
&Trees &Bikes &Ubc &Leuven &Wall  &Graffiti   \\

\midrule
Harris& 0.342  & 0.728 & 0.746 & 0.760 & 0.521 & 0.254\\

Harris-Laplace & 0.221  & 0.332 & 0.685 & 0.578 & 0.409 & 0.319\\

FOAGDD & 0.731  & \textbf{0.880} & 0.856 & \textbf{0.900} & 0.562 &0.269 \\
\hline
SOGGDD& \textbf{0.832} & 0.877 & 0.844 & 0.854 &\textbf{0.659}& 0.299\\

DoG & 0.405  & 0.566 & 0.691 & 0.668 & 0.412 & 0.320\\

SURF & 0.487  & 0.812 & 0.776 & 0.725 & 0.489 & 0.304\\

KAZE & 0.569  & 0.744 & \textbf{0.885} & 0.671 & 0.612 & \textbf{0.362}\\
\hline
FAST & 0.378 & 0.639 & 0.776 & 0.707 & 0.429 & 0.266  \\
\hline

ACJ & 0.459  & 0.586 & 0.510 & 0.642 & 0.434 & 0.277\\

ANDD & 0.308  & 0.428 & 0.540 & 0.600 & 0.214 & 0.230\\
CPDA & 0.459  & 0.479 & 0.590 & 0.558 & 0.316 & 0.205\\
GCM & 0.429  & 0.550 & 0.601 & 0.613 & 0.302 & 0.243\\

New-Curvature & 0.471  & 0.527 & 0.645 & 0.664 & 0.315 & 0.213\\

He \& Yung & 0.394  & 0.617 & 0.644 & 0.556 & 0.314 & 0.250\\
\hline
D2-Net & 0.561  & 0.658 & 0.727 & 0.681 & 0.530 & 0.185\\

LF-Net & 0.640  & 0.741 & 0.796 & 0.742 & 0.558 & 0.277\\

LIFT & 0.368  & 0.418 & 0.667 & 0.573 & 0.436 & 0.211\\

DT-CovDet & 0.715  & 0.820 & 0.789 & 0.783 & 0.587 & 0.237\\
\bottomrule
\end{tabular}}
\end{table}

\subsubsection{Repeatability under affine transformation}
In this experiment, one thousand images~\cite{deng2009imagenet} which contain different types of scenes without ground truths are utilized to measure the average repeatabilities for the eighteen methods. With affine transformation, JPEG compression, and noise degradation of each original image, a total of 217,000 transformed test images are obtained as follows:

$\bullet$ Rotation: An input image is rotated with an interval of $\pi/18$ within the range of $[-\pi/2,\pi/2]$, not including 0.

$\bullet$ Uniform scaling: The scale factors $s_x=s_y$ are in $[0.5,2]$ with an interval of 0.1, not including 1.

$\bullet$ Non-uniform scaling: The scale $s_x$ is in $[0.7,1.5]$ and $s_y$ is in $[0.5,1.8]$ with an interval of 0.1, not including $s_x=s_y$.

$\bullet$ Shear transformations: The shear factor $f$ is in $[-1,1]$ with an interval of 0.1, not including 0, with the following equation
\begin{equation}\begin{aligned}
\left [ \begin{array}{c}
m'\\
n'\end{array} \right]= \left [ \begin{array}{cc}
1&f\\
0&1
\end{array} \right]\left [ \begin{array}{c}
m\\
n\end{array} \right].\nonumber\end{aligned}\end{equation}

$\bullet$ Lossy JPEG compression: The compression factor is in $[5,100]$ with an interval of 5.

$\bullet$ Gaussian noise: Zero-mean white Gaussian noise is added to the input image at 15 variances in $[1,15]$ with an interval of 1.\\

\begin{figure*}[!t]
\setlength{\abovecaptionskip}{-2pt}
\setlength{\belowcaptionskip}{0pt}
\centering
\includegraphics[width=6.5in]{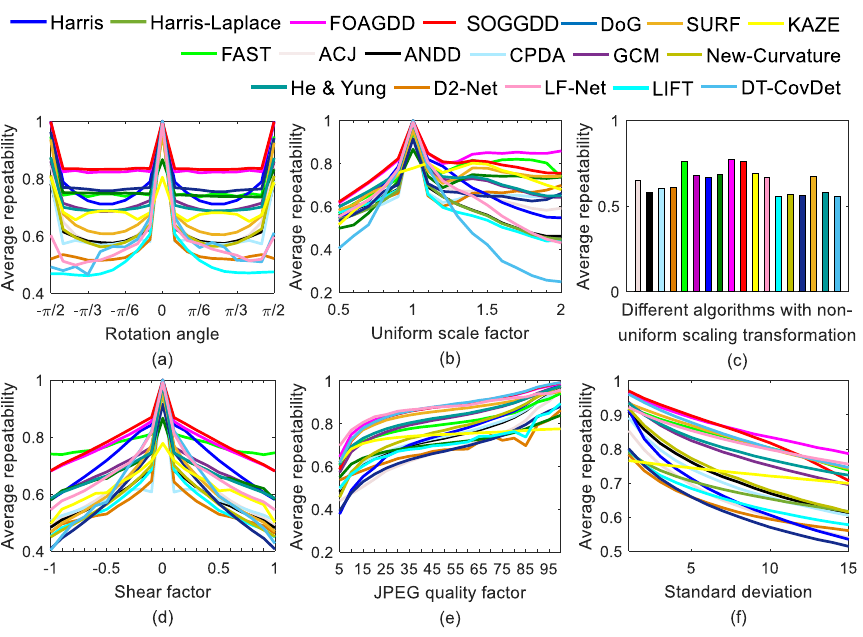}
\caption{Average repeatabilities of the eighteen detectors under rotation, uniform scaling, non-uniform scaling, shear transforms, lossy JPEG compression, and zero-mean white Gaussian noises.}
\label{fig9}
\end{figure*}

\indent With this criteria, the performances of the eighteen methods with their default tuneable parameters are evaluated. In this experiment, if an interest point is extracted in a transformed image, and it is near the ground truth location (say within 4 pixels), then a repeated interest point is extracted. The average repeatabilities with different rotation, uniform scaling, non-uniform scaling, shear transformation, lossy JPEG compression, and noise degradation of the eighteen methods are shown in Fig.~\ref{fig9}. It can be found that the FOAGDD method~\cite{zhang2020corner} and the SOGGDD method~\cite{zhang2019corner2} achieved the best and the second best detection performance in this evaluation criteria respectively.

\renewcommand{\arraystretch}{1.3}
\begin{table}[!t]
\setlength{\abovecaptionskip}{0pt}
\setlength{\belowcaptionskip}{0pt}
\footnotesize

\caption{Average results for repeatability under affine transformations.}
\label{test4}
\centering
\setlength{\tabcolsep}{1.2mm}{
\begin{tabular}{ccccccc}

\toprule
\multirow{3}{*}{Detector} & \multirow{3}{*}{Rotation}  & \multirow{2}{*}{Uniform}  & \multirow{1}{*}{Non-}  & \multirow{3}{*}{Shear} & \multirow{3}{*}{JPEG}  & \multirow{2}{*}{Gaussian }  \\
~ & ~  & \multirow{2}{*}{scale} & \multirow{1}{*}{uniform} & ~ &  & \multirow{2}{*}{noise} \\
~ & ~  &  & \multirow{1}{*}{scale} & ~ &  & ~ \\
\midrule
Harris& 0.784 & 0.674 & 0.665 & 0.722 & 0.742 & 0.666 \\

Harris-& \multirow{2}{*}{0.762} & \multirow{2}{*}{0.695} & \multirow{2}{*}{0.688} & \multirow{2}{*}{0.671} & \multirow{2}{*}{0.711} & \multirow{2}{*}{0.681} \\
Laplace & ~  &  & ~ & ~ &  & ~ \\
FOAGDD & 0.851 & \textbf{0.810} & \textbf{0.774} & 0.775 & \textbf{0.875} & \textbf{0.862} \\
\hline
SOGGDD& \textbf{0.860} & 0.782 & 0.761 & \textbf{0.784} & 0.869 & 0.845 \\

DoG & 0.794 & 0.657 & 0.561 & 0.580 & 0.673 & 0.619 \\

SURF & 0.681 & 0.733 & 0.673 & 0.613 & 0.850 & 0.831 \\

KAZE & 0.693 & 0.719 & 0.692 & 0.621 & 0.740 & 0.731 \\
\hline
FAST & 0.775 & 0.757 & 0.761 & 0.776 & 0.799 & 0.826 \\
\hline
ACJ & 0.648 & 0.662 & 0.648 & 0.622 & 0.682 & 0.653 \\

ANDD & 0.635 & 0.599 & 0.578 & 0.595 & 0.743 & 0.720 \\

CPDA & 0.621 & 0.652& 0.602 & 0.590 & 0.748 & 0.706 \\

GCM & 0.732 & 0.708 & 0.678 & 0.687 & 0.822 & 0.790 \\

New- & \multirow{2}{*}{0.637} & \multirow{2}{*}{0.592} & \multirow{2}{*}{0.570} & \multirow{2}{*}{0.582} & \multirow{2}{*}{0.769} & \multirow{2}{*}{0.730} \\
Curvature & ~  &  & ~ & ~ &  & ~ \\
He \& Yung & 0.728 & 0.714 & 0.578 & 0.677 & 0.830 & 0.803 \\
\hline
D2-Net & 0.556 & 0.665 & 0.609 & 0.561 & 0.681 & 0.633 \\

LF-Net & 0.577 & 0.632 & 0.670 & 0.671 & 0.873 & 0.825 \\

LIFT & 0.525 & 0.583 & 0.556 & 0.572 & 0.705 & 0.655 \\

DT-CovDet & 0.577 & 0.509 & 0.559 & 0.617 & 0.873 & 0.838 \\
\bottomrule
\end{tabular}}
\end{table}
\subsubsection{Average matching scores under the VLBenchmarks}
In this experiment, the HPatches image dateset~\cite{balntas2017hpatches} which contains 116 scenes (58 scenes for viewpoint transformations, 58 scenes for lighting changes, and each scene having 6 images) are utilized to measure the average matching scores for the eighteen methods with four different feature descriptors (i.e., BRIEF~\cite{calonder2010brief}, SIFT~\cite{lowe2004distinctive}, Hard-Net++~\cite{HardNet2017}, and L2-Net~\cite{tian2017l2}). In this evaluation, the threshold for each detected method is selected so that each method detects approximately 1,000 interest points from each test image. The code for image matching is provided by VLBenchmarks~\cite{lenc12vlbenchmarks}.

\begin{figure*}[!t]
\setlength{\abovecaptionskip}{2pt}
\setlength{\belowcaptionskip}{2pt}
\centering
\includegraphics[width=7.2in]{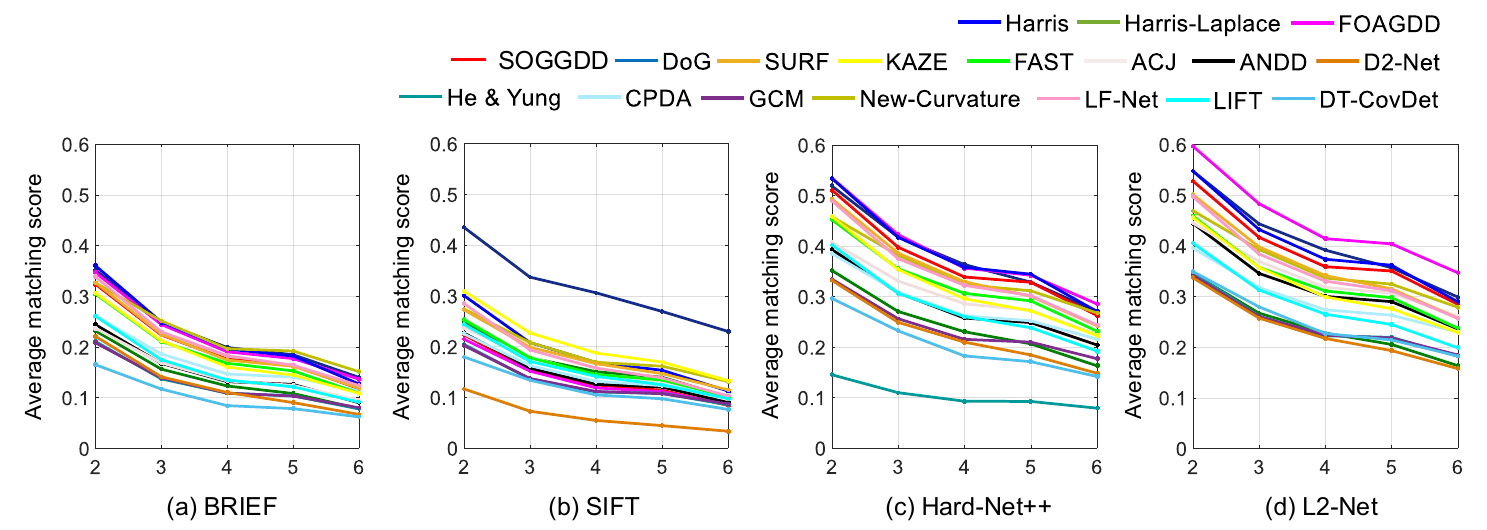}
\caption{Average matching scores of the eighteen detection methods with BRIEF~\cite{calonder2010brief}, SIFT~\cite{lowe2004distinctive}, Hard-Net++~\cite{HardNet2017}, and L2-Net~\cite{tian2017l2}.}
\label{figms}
\end{figure*}

The average matching scores with different lighting conditions and viewpoints of the eighteen methods are shown in Fig.~\ref{figms} and Table.~\ref{match}. It can be found from Fig.~\ref{figms} that choosing different descriptors has an impact on the performance of matching scores of the eighteen methods. It can also be found that the eighteen methods with the L2-Net descriptor~\cite{tian2017l2} achieve the best performance for the HPatches image dateset. From Fig.~\ref{figms}(d), it is observed that the FOAGDD method~\cite{zhang2020corner} and the DoG method~\cite{lowe2004distinctive} achieved the best and the second matching score performance respectively under this evaluation criteria.
\begin{table}[htbp]
\setlength{\abovecaptionskip}{0pt}
\setlength{\belowcaptionskip}{0pt}
\footnotesize
\renewcommand{\arraystretch}{1.5}
\caption{Average match percentage.}
\label{match}
\centering
\begin{tabular}{ccccc}

\hline
\multirow{2}{*}{Detector} & \multicolumn{4}{c}{Descriptor}  \\
\cline{2-5}
~~& Brief  & SIFT  & Hard-Net++&L2-Net   \\
\midrule[1pt]
Harris & 0.223 & 0.189 & 0.385 &0.401 \\

GCM & 0.128 & 0.129 &0.238 &0.246  \\

FAST & 0.189 & 0.162&0.327  &0.332  \\

D2-Net &0.126  & 0.064 & 0.224 &0.232 \\

CPDA & 0.169 & 0.149 &0.284&0.296 \\

ANDD & 0.152 & 0.144 &0.282 &0.323 \\

ACJ & 0.151 & 0.159 &0.304 & 0.340\\

Harris-Laplace &0.140  &0.164  &0.244 &0.242 \\

FOAGDD & 0.219 & 0.136 &\textbf{0.388} & \textbf{0.449} \\

SOGGDD& 0.201 & 0.140 &0.367 & 0.387\\

KAZE & 0.186& 0.205 &0.320 &0.324 \\

LF-Net &0.208  &0.175  &0.346 & 0.356\\

LIFT &0.156  & 0.155 &0.280 &0.285 \\

New-Curvature & 0.223 & 0.189 &0.348 &0.360 \\

DoG &\textbf{0.225}  &\textbf{0.315}  &0.380 &0.407 \\

SURF & 0.201 & 0.180 &0.350 & 0.362\\

He \& Yung & 0.127& 0.129 & 0.104&0.246\\
DT-CovDet&0.101 &0.118 &0.205 & 0.250   \\

\hline

\end{tabular}
\end{table}

\subsection{Execution time and memory usage}
Six test images~\cite{aanaes2012interesting} (`House', `Book ', `House ', `Teddy Bear ', `Greens ', and `Groceries') are used in this experiment. Each interest point detector has been implemented in MATLAB (R2018b) using a 2.20 GHz CPU with 128 GB of memory. The GPU model is RTX1080ti with 12 GB video memory. For each test image, an interest point detector was executed 100 times and the average execution time and memory were measured. Table~\ref{testteim} shows the execution time and memory usage comparisons on the eighteen methods for image sizes of 1,200 $\times$ 1,600 for detecting around 1,000 features. It can be found from Table~\ref{testteim} that the FAST method~\cite{rosten2008faster} has the shortest execution time and the Harris method~\cite{harris1988combined} has the least memory usage in this evaluation criteria.

\begin{table*}[!htbp]
\setlength{\abovecaptionskip}{0pt}
\setlength{\belowcaptionskip}{0pt}
\footnotesize
\renewcommand{\arraystretch}{1.5}
\caption{Comparisons on execution time and memory usage (image size is in pixels. execution time and memory usage are in second and MB respectively).}
\label{testteim}
\centering
\begin{tabular}{ccccccccccccc}
\toprule
\multirow{5}{*}{Detector} &\multicolumn{2}{c}{House (Set 1)}   & \multicolumn{2}{c}{Book (Set 2)} & \multicolumn{2}{c}{House (Set 8)}&\multicolumn{2}{c}{Teddy Bear (Set 9)}  & \multicolumn{2}{c}{Greens (Set 26)}& \multicolumn{2}{c}{Groceries (Set 30)}\\
~&\multicolumn{2}{c}{($1200 \times 1600$)}   & \multicolumn{2}{c}{($1200 \times 1600$)} &\multicolumn{2}{c}{($1200 \times 1600$)}&\multicolumn{2}{c}{($1200 \times 1600$)}   & \multicolumn{2}{c}{($1200 \times 1600$)} &\multicolumn{2}{c}{($1200 \times 1600$)}\\
\cline{2-13}
~&\multicolumn{1}{c}{Executi-} & \multicolumn{1}{l}{Memory}    & \multicolumn{1}{c}{Executi-} & \multicolumn{1}{l}{Memory} &\multicolumn{1}{c}{Executi-}  & \multicolumn{1}{l}{Memory}    & \multicolumn{1}{c}{Executi-}  & \multicolumn{1}{l}{Memory}  &\multicolumn{1}{c}{Executi-}  & \multicolumn{1}{l}{Memory} &\multicolumn{1}{c}{Executi-}  & \multicolumn{1}{l}{Memory}\\
~& \multicolumn{1}{c}{on time} & \multicolumn{1}{c}{usage}  &  \multicolumn{1}{c}{on time}& \multicolumn{1}{c}{usage}&\multicolumn{1}{c}{on time}& \multicolumn{1}{c}{usage}   & \multicolumn{1}{c}{on time}& \multicolumn{1}{c}{usage} &\multicolumn{1}{c}{on time}& \multicolumn{1}{c}{usage} &\multicolumn{1}{c}{on time}& \multicolumn{1}{c}{usage}\\
~& \multicolumn{1}{c}{(s)} & \multicolumn{1}{c}{(MB)}  &  \multicolumn{1}{c}{ (s)}& \multicolumn{1}{c}{(MB)}&\multicolumn{1}{c}{(s)}& \multicolumn{1}{c}{(MB)}   & \multicolumn{1}{c}{(s)}& \multicolumn{1}{c}{(MB)} &\multicolumn{1}{c}{(s)}& \multicolumn{1}{c}{(MB)} &\multicolumn{1}{c}{(s)}& \multicolumn{1}{c}{(MB)}\\
\hline
Harris  & 0.420&  \textbf{205.9} & 0.422  & \textbf{201.1}&0.427  & \textbf{207.5}& 0.433&  \textbf{181.3}&0.436&  \textbf{178.0} &0.431 &  \textbf{185.2} \\

Harris-  & \multirow{2}{*}{1.994} & \multirow{2}{*}{295.6}& \multirow{2}{*}{2.042} & \multirow{2}{*}{234.4}& \multirow{2}{*}{1.997} & \multirow{2}{*}{220.3}&\multirow{2}{*}{2.386} & \multirow{2}{*}{236.6}& \multirow{2}{*}{2.282} & \multirow{2}{*}{204.8}& \multirow{2}{*}{2.165} & \multirow{2}{*}{208.9}\\
Laplace  & & & & & & \\
FOAGDD  & 39.176 & 373.2& 39.431 & 282.7& 43.639 &320.1& 39.578 &287.3& 48.140 & 305.8& 38.599 & 277.0\\
\hline
SOGGDD &23.028 & 355.7 &24.808 & 276.8& 38.465 & 303.0&32.304 & 266.8& 31.790 & 264.0&27.777 & 297.6  \\

DoG & 0.520 & 215.9&0.517 & 209.7 & 0.519 & 231.4 & 0.538 & 200.1&0.518 & 235.0& 0.519 & 178.2\\

SURF  & 0.133 & 298.1&0.108 & 277.7 &0.116 & 236.2& 0.114 & 221.7&0.110 & 244.2 &0.110 & 236.2 \\

KAZE  & 2.003 & 307.0& 1.405 & 247.5&1.502 & 283.6 &1.669 & 225.7& 1.284 & 270.3&1.708 & 237.1 \\
\hline
FAST   &\textbf{0.018} & 340.3 &\textbf{0.009} & 248.7 & \textbf{0.009} & 229.5 &\textbf{0.015} & 226.8 &\textbf{0.009} & 205.9& \textbf{0.012} & 218.8 \\
\hline

GCM & 4.537 & 378.6&2.370 & 305.4 &2.704 & 271.8 & 2.561 & 288.5&2.313 & 267.1 &4.162 & 284.1\\

New-&\multirow{2}{*}{4.674} & \multirow{2}{*}{329.2}&\multirow{2}{*}{2.443} &\multirow{2}{*}{ 255.9} &\multirow{2}{*}{2.774} &\multirow{2}{*}{ 254.8}&\multirow{2}{*}{ 2.573} & \multirow{2}{*}{292.6}&\multirow{2}{*}{2.324} & \multirow{2}{*}{264.0} &\multirow{2}{*}{4.262} & \multirow{2}{*}{261.3}  \\
Curvature & & & & & & \\
CPDA & 5.177 &  262.9&2.899 &  274.1 &3.632 &  248.4 & 3.867 &  258.8&4.016 &  237.2 &4.902 &  259.3 \\

ANDD  & 20.203 &  372.3&10.412 &  352.8 &10.799 & 360.4 & 7.770 &  346.4&5.449 &  283.0 &17.899 & 336.1\\

ACJ & 23.486 & 342.1&13.650 &  418.2 &13.973 &  373.3 & 25.893 &  432.4&13.303 &  391.1&18.107 &  396.8 \\

He \&  & \multirow{2}{*}{5.052} &  \multirow{2}{*}{338.2}&\multirow{2}{*}{2.890} &  \multirow{2}{*}{292.0} &\multirow{2}{*}{3.585} &  \multirow{2}{*}{271.3} & \multirow{2}{*}{3.920} &  \multirow{2}{*}{251.0}&\multirow{2}{*}{4.098}& \multirow{2}{*}{ 252.9} &\multirow{2}{*}{4.748} &  \multirow{2}{*}{282.4}\\
 Yung & & & &   & &  &  &  & &  & & \\
\hline
D2-Net  & 5.185 & 1508.9&5.167 & 1507.7 &5.218 & 1514.0 & 5.231 & 1513.6&5.206 & 1507.2 &5.123 & 1515.4\\

LF-Net & 0.245 & 1569.4&0.243 &1443.7 &0.242 & 1442.0 & 0.244 & 1441.9&0.245 & 1441.9&0.247 & 1569.9\\

LIFT  & 35.086 & 590.5&35.878 & 590.4&35.667 & 589.1 & 35.116 & 590.7&34.763 & 589.7 &34.024 & 589.9  \\

DT-& \multirow{2}{*}{0.404} & \multirow{2}{*}{1493.8}&\multirow{2}{*}{0.400} & \multirow{2}{*}{1491.7} &\multirow{2}{*}{0.367} & \multirow{2}{*}{1488.7} &\multirow{2}{*}{0.408} & \multirow{2}{*}{1491.3}&\multirow{2}{*}{0.426} & \multirow{2}{*}{1488.9} &\multirow{2}{*}{0.395} & \multirow{2}{*}{1494.4} \\
CovDet& & & &   & &  &  &  & &  & & \\
\bottomrule

\end{tabular}
\end{table*}

\subsection{Overview of IFI extraction techniques for interest point detection}
According to the experimental results of the eighteen state-of-the-art methods, we will overview the different types of IFI extraction techniques for interest point detection in this subsection.

The experimental results for the eighteen state-of-the-art methods on the three evaluation criteria in~\cite{aanaes2012interesting, awrangjeb2012performance, lenc12vlbenchmarks} are illustrated in Fig.~\ref{fig14}, Fig.~\ref{fig9}, and Fig.~\ref{figms}. From the two evaluation criteria in~\cite{aanaes2012interesting, awrangjeb2012performance}, it can be observed that the FOAGDD~\cite{zhang2020corner}, SOGGDD~\cite{zhang2019corner2}, and FAST methods~\cite{rosten2008faster} achieves the best, second, and third detection performances. The main reason is that compared with other algorithms, these three methods~\cite{zhang2020corner, zhang2019corner2, rosten2008faster} use multi-directional filtering instead of filtering along the horizontal and vertical directions. The use of multi-directional filtering can extract more accurate IFI for interest point detection. Meanwhile, it can also be found that the detection performances of the other intensity based methods (Harris~\cite{harris1988combined}, Harris-Laplace~\cite{mikolajczyk2004scale}, DoG~\cite{lowe2004distinctive}, SURF~\cite{bay2006surf}, and KAZE~\cite{alcantarilla2012kaze}) are relatively poor. The reason is that the authors of above five methods have not considered how to accurately extract IFI. The first or second-order derivative extraction along the orthogonal directions cannot accurately describe the IFI of an interest point~\cite{zhang2020corner, zhang2019corner2}. Take a blob type interest point model as an example as illustrated in Fig.~\ref{blobb}(a), its corresponding SOIGDD is shown in the second column of Fig.~\ref{blobb}. It can be observed that the two orthogonal SOIGDDs do not contain enough IFI for the blob type interest point. After the blob interest point undergoes an affine image transformation as illustrated in Fig.~\ref{blobb}(b), its corresponding SOIGDD is illustrated in the second column of Fig.~\ref{blobb}. It can also be found that the two orthogonal SOIGDDs shown in Fig.~\ref{blobb} do not contain enough IFI for the blob type interest point. Furthermore, there is a large difference on the SOIGDDs between the original blob and the blob undergoing an affine image transformation. Then the blob interest point may not be properly extracted under such or similar affine image transformations. From the evaluation criteria in~\cite{lenc12vlbenchmarks}, it can be found that using different descriptors have an impact on the performance of matching scores of the eighteen methods. Fig.~\ref{figms} shows that the eighteen methods with the L2-Net descriptor~\cite{tian2017l2} achieve the best performance in the HPatches image dateset. It can also be found from Fig.~\ref{figms}(d) that the FOAGDD method~\cite{zhang2020corner} and the DoG method~\cite{lowe2004distinctive} achieved the best and the second best matching score performances in this evaluation criteria respectively.

Two other performance measures for interest point detection are execution time and memory usage. As illustrated in Table~\ref{testteim}, the FAST algorithm~\cite{rosten2008faster} has the shortest execution time among the eighteen methods. It can also be observed that the SURF, LF-Net, DT-CovDet, Harris, and DoG methods~\cite{bay2006surf, ono2018lf, zhang2017learning, harris1988combined, lowe2004distinctive} are excellent with performance indicators on execution time. It is worth to note that the ANDD, ACJ, SOGGDD, LIFT, and FOAGDD methods~\cite{shui2013corner, xia2014accurate, zhang2019corner2, yi2016lift, zhang2020corner} cannot meet the needs of real-time applications. They can be implemented using GPU~\cite{Cornelis2008Fast1} or FPGA~\cite{huang2012high} to improve the speed performance. Meanwhile, the Harris method~\cite{harris1988combined} has the least memory usage among the eighteen methods as illustrated in Table~\ref{testteim}. The memory usages for the intensity based methods and edge based methods do not have significant differences. Furthermore, it can be found that the memory usages for machine learning based methods (LIFT~\cite{yi2016lift}, D2-Net~\cite{dusmanu2019d2}, LF-Net~\cite{ono2018lf}, and DT-CovDet~\cite{zhang2017learning}) are much higher than other algorithms.

\section{Some future research directions for interest point detection}
Although considerable progresses have been made in theory and performance, image interest point detection is still an open problem and faces a range of challenges.

\begin{figure}[!htbp]
\setlength{\abovecaptionskip}{5pt}
\setlength{\belowcaptionskip}{0pt}
\centering
\includegraphics[width=3.5in]{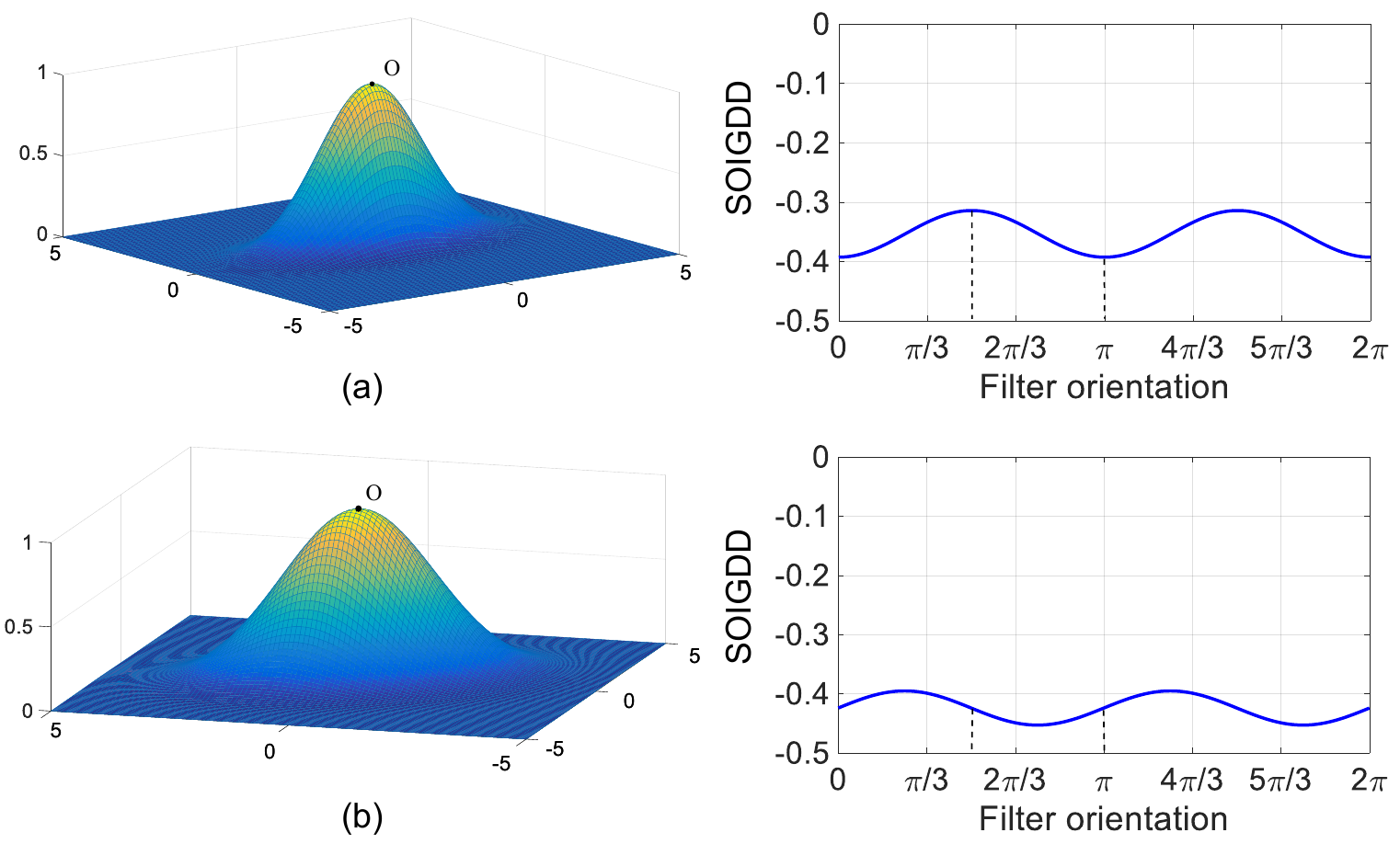}
\caption{Examples of SOIGDD changes with image affine image transformation. (a) a blob-type interest point, (b) the blob-type interest point undergoing an affine image transformation.}
\label{blobb}
\end{figure}

$\bullet$ Multi-directional IFI extraction with multiple scales: We show that most multi-scale IFI extraction techniques focus on the multi-scale IFI extraction along the two orthogonal directions. As shown in Fig.~\ref{blobb}, the existing multi-scale IFI extraction approaches along the orthogonal directions cannot properly extract IFI. It is necessary for us to extract IFI from an input image along multi-directions with multiple scales. Up to now, no one has proposed a theoretical framework about how to accurately extract multi-directional IFI with multiple scales for better detecting interest points.

\begin{figure}[!htbp]
\setlength{\abovecaptionskip}{0pt}
\setlength{\belowcaptionskip}{0pt}
\centering
\includegraphics[width=3.3in]{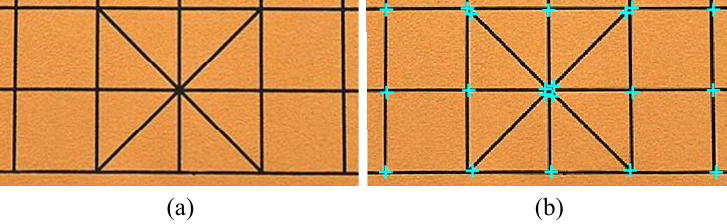}
\caption{Examples of the adjacent interest points detection results by an interest point detector. (a) a local Chinese chessboard image, (b) the adjacent interest points detection results from the FOAGDD method.}
\label{adjacent}
\end{figure}

$\bullet$ IFI extraction for adjacent interest points detection: The existing IFI extraction approaches for interest point detection (especially for intensity variation based IFI extraction) have not considered how to accurately extract the IFI for adjacent interest points. Take a local region of a Chinese chessboard image as an example, the interest point detection results from the FOAGDD method~\cite{zhang2020corner} for this region are illustrated in Fig.~\ref{adjacent} (b). It can be found that the adjacent interest points cannot be accurately detected by the FOAGDD method. The reason is that there are mutual influences for the IFIs between the adjacent interest points. The mutual influences of the IFIs lead to the existing interest point detection methods not accurately detecting the adjacent interest points, or only detecting one of the interest points, or the localizations of the detected interest points being seriously offset as illustrated in Fig.~\ref{adjacent} (b).

\begin{figure}[!htbp]
\setlength{\abovecaptionskip}{0pt}
\setlength{\belowcaptionskip}{0pt}
\centering
\includegraphics[width=3.3in]{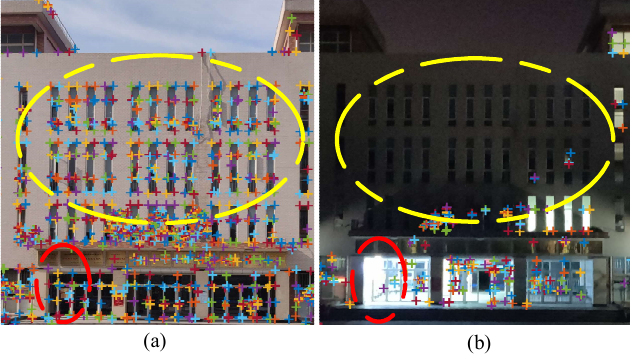}
\caption{Examples of the interest point detection results by an interest point detector under different light conditions. }
\label{figlight}
\end{figure}

$\bullet$ IFI extraction for interest point detection under different imaging conditions: The existing interest point detection methods have not accurately considered how to determine interest points from the same scene under different imaging conditions. Take a building scene as an example, the interest point extraction results from the FOAGDD detector~\cite{zhang2020corner} under different illuminations are illustrated in Fig.~\ref{figlight}. It can be found that some obvious interest points in the window area (marked by `$\LARGE\rotatebox{90} o$') cannot be extracted under different illuminations as illustrated in Fig.~\ref{figlight}(a) and (b). The reason is that the magnitude information of the IFI for the interest point at the same position will change with the change of imaging conditions.

\begin{figure}[h!]
\setlength{\abovecaptionskip}{0pt}
\setlength{\belowcaptionskip}{0pt}
\centering
\includegraphics[width=3.5in]{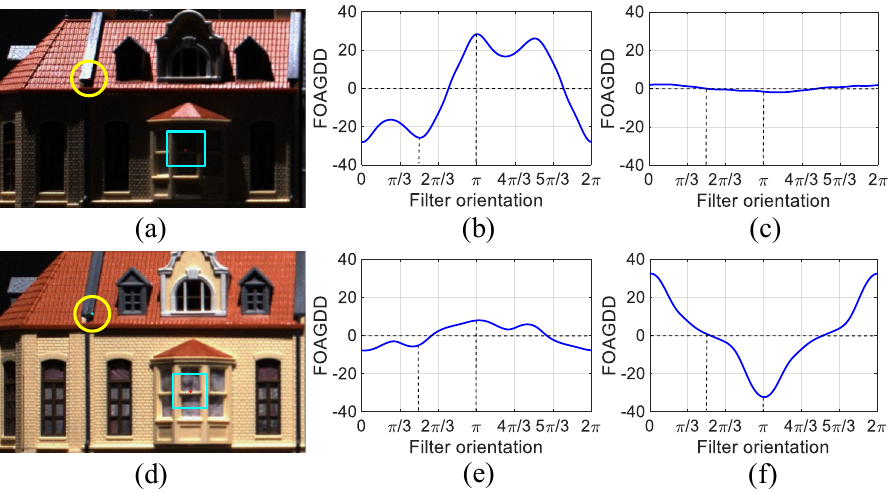}
\caption{Examples of the FOAGDDs at an interest point and an edge point at the same locations under different imaging conditions. (a) and (d) are the images with different lighting condition, (b) and (e) are corresponding the FOAGDD responses with multi-directions for an interest point (marked by `$\bigcirc$'), and (c) and (f) are corresponding the FOAGDD responses with multi-directions for an edge point (marked by `$\Box$'). }
\label{GBD}
\end{figure}

Furthermore, our research also indicates that different imaging conditions will cause the first-order derivatives at edge points to be larger than that at interest points in many directions. Take the  test image `House' under different imaging condition as an example, an interest point and an edge point at the same location in Fig.~\ref{GBD}(a) and (d) are marked by `$\bigcirc$' and `$\Box$' respectively. It can be seen from Fig.~\ref{GBD}(b) and (c) that the corresponding FOAGDD of the interest point is larger than the FOAGDD of the edge point in most directions. Then the FOAGDD method~\cite{zhang2020corner} can easily detect the interest point in the scene in Fig.~\ref{GBD}(a). If the scene undergoes a different imaging condition as shown in Fig.~\ref{GBD}(d). It can be seen from Fig.~\ref{GBD}(e) and (f) that the FOAGDD for the interest point is less than the FOAGDD for the edge point in most directions. Then the interest point may not be detected with such or similar imaging conditions by the FOAGDD detector~\cite{zhang2020corner}. Therefore, in order to better detect interest points from an input image, in addition to the amplitude information of the intensity variations of the image pixels, it is necessary to combine other techniques (e.g., image illumination robust strategy~\cite{gevrekci2009illumination} and the difference between the waveforms of the intensity variations at the interest points and the step edges~\cite{zhang2020corner}) to extract and analyze the IFI for image pixels.

\begin{figure}[h!]
\setlength{\abovecaptionskip}{0pt}
\setlength{\belowcaptionskip}{0pt}
\centering
\includegraphics[width=3.5in]{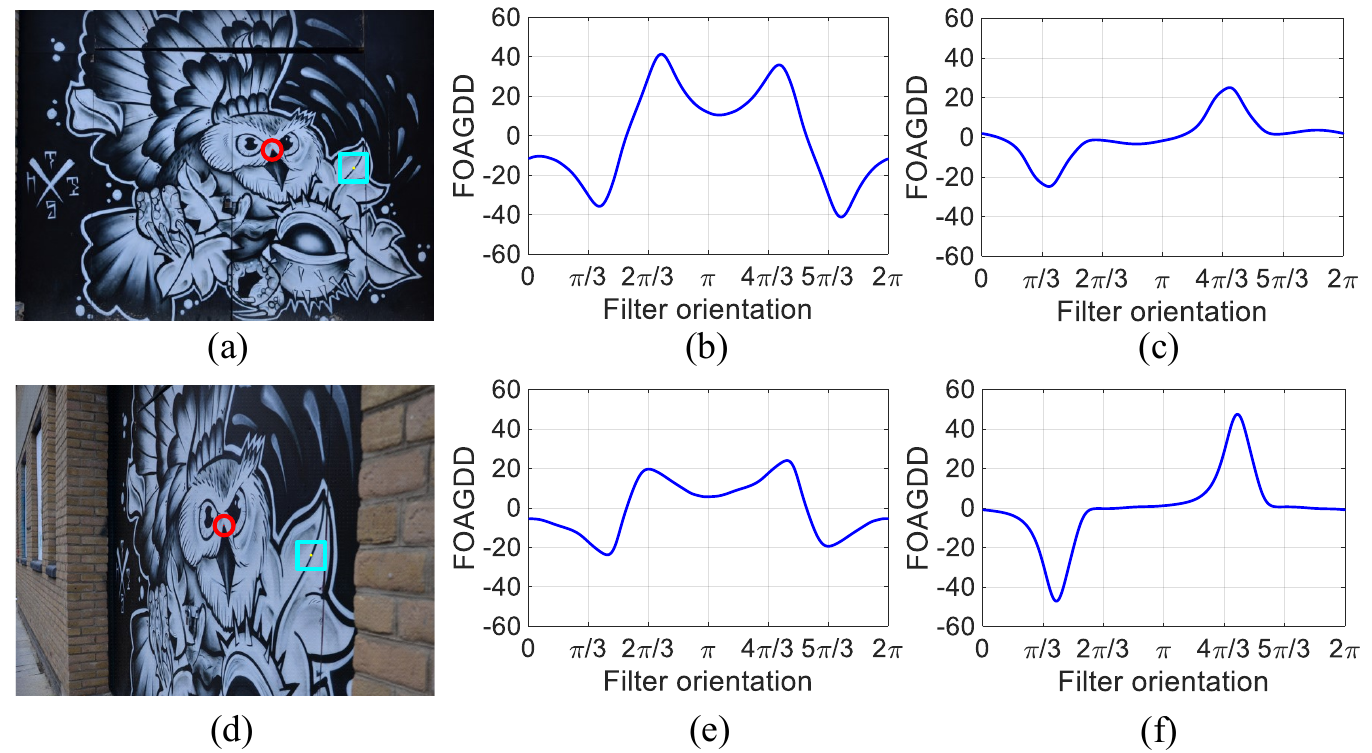}
\caption{Examples of the FOAGDDs at an interest point and an edge point at the same location under image affine transformations. (a) and (d) are the images under different affine transformations, (b) and (e) are corresponding the FOAGDD responses with multi-directions for an interest point (marked by `$\bigcirc$'), and (c) and (f) are corresponding the FOAGDD responses with multi-directions for an edge point (marked by `$\Box$').}
\label{figaffine}
\end{figure}

$\bullet$ IFI extraction for interest point detection under image affine transformations: The existing interest point detection methods have not considered how to extract interest points from an input image under affine transformations. Take the test image `v\_bird' as an example as shown in Fig.~\ref{figaffine}(a), an interest point and an edge point at the same location in Fig.~\ref{figaffine}(a) and (d) are marked by `$\bigcirc$' and `$\Box$' respectively. As illustrated in Fig.~\ref{figaffine}(b) and (c), the corresponding FOAGDD for the interest point is larger than the FOAGDD at the edge point in most directions. Then the interest point can be easily detected by the FOAGDD method~\cite{zhang2020corner} from the original image. If the original image undergoes an affine transformation as illustrated in Fig.~\ref{figaffine}(d). As illustrated in Fig.~\ref{figaffine}(e) and (f), the corresponding FOAGDD for the interest point is less than the FOAGDD for the edge point in many directions. Then the interest point may not be detected with such or similar image affine transformations. Within the scope of our investigations, no one has proposed how to accurately extract IFI for accurately detecting interest points under the conditions of image affine transformations.

$\bullet$ Deep learning techniques for interest point detection: In recent years, deep learning techniques have shown tremendous improvements in many computer vision tasks such as image segmentation and image classification. However, the four deep learning based methods (LIFT~\cite{yi2016lift}, D2-Net~\cite{dusmanu2019d2}, LF-Net~\cite{ono2018lf}, and DT-CovDet~\cite{zhang2017learning}) for interest point detection do not have the advantage on detection and image matching performances over the traditional intensity based methods on the three performance evaluation criteria (i.e., average repeatabilities under image affine transformation~\cite{awrangjeb2008robust}, average recall rates on the DTU-Robots dataset~\cite{aanaes2012interesting}, and average matching score under the VLBenchmarks~\cite{lenc12vlbenchmarks}). Thus, the potential capacity of deep learning approaches for accurate interest point detection can be further explored in the future.

$\bullet$  The application of interest point detection: Image interest point detection is one of the fundamental tasks in image analysis and computer vision. However, the application of interest point detection in computer vision has not been fully explored. Therefore, a promising research direction is to customize modern image interest point detection techniques to meet the different requirements of actual computer vision tasks such as structure-from-motion, object tracking, and image-guided robotic-assisted surgery.

\section{Conclusion}
As a contemporary interest point detection survey, this paper presented a taxonomy of the image feature information extraction techniques for interest point detection, discussed advantages and disadvantages of the existing interest point detection methods, introduced the existing popular datasets and evaluation criteria, and analyzed the performances for the eighteen most representative interest point detection methods in terms of average recall rates under different camera positions and lighting changes, average repeatabilities under different image affine transformations, JPEG compressions, and noise degradations, and matching scores based on the VLBenchmarks with four different descriptors. Furthermore, the development trend for image feature information extraction for interest point detection is elaborated. We hope that this survey will not only enable researchers to better understand image interest point detection but also inspire future research activities.

\appendices
\section*{Acknowledgment}
This work was supported by the Youth Innovation Team of Shaanxi Universities.
\ifCLASSOPTIONcaptionsoff
  \newpage
\fi

\bibliographystyle{IEEEtran}
\bibliography{arxiv_IFI}

\begin{IEEEbiography}[{\includegraphics[width=1in,height=1.25in,clip,keepaspectratio]{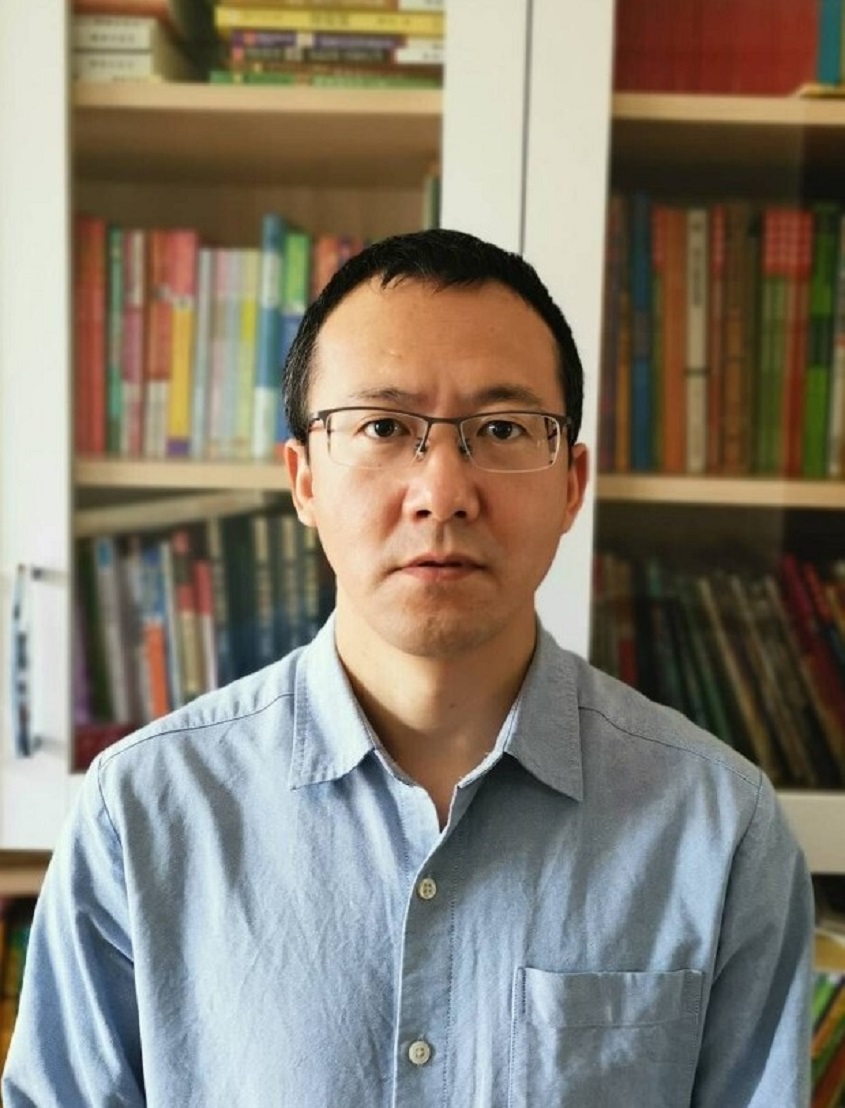}}]{Junfeng Jing}
received the MS degree in electrical engineering from from Xi'an Polytechnic University in China and the PhD degree in mechatronic engineering from Xidian University in China. He is a professor at Xi'an Polytechnic University, China. His research interests include artificial intelligence, machine vision, image processing, pattern recognition, and industrial robot control.
\end{IEEEbiography}
\begin{IEEEbiography}[{\includegraphics[width=1in,height=1.25in,clip,keepaspectratio]{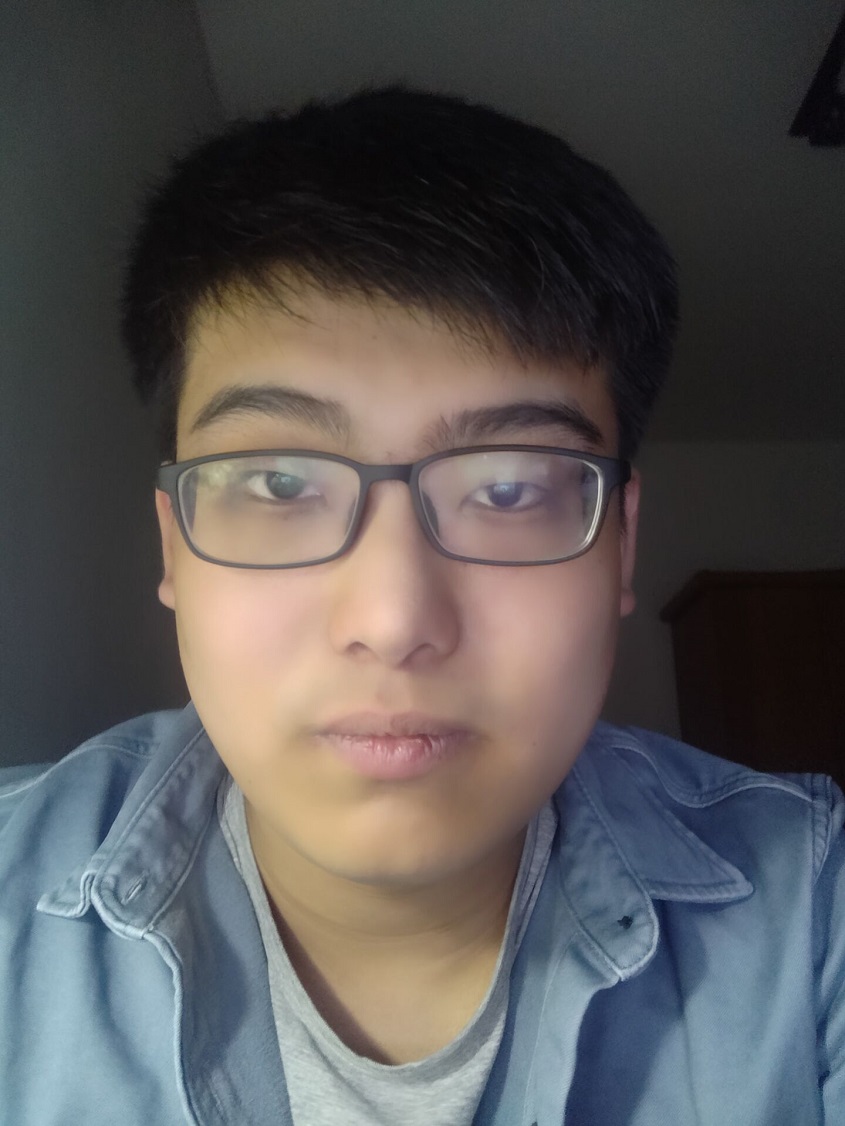}}]{Tian Gao}
received the BS degree from the Xi'an University of Technology in China and he is pursuing the Master degree with the school of electronic and information, Xi'an Polytechnic University, China. His research interests include computer vision, image processing, and machine learning.
\end{IEEEbiography}
\begin{IEEEbiography}[{\includegraphics[width=1in,height=1.25in,clip,keepaspectratio]{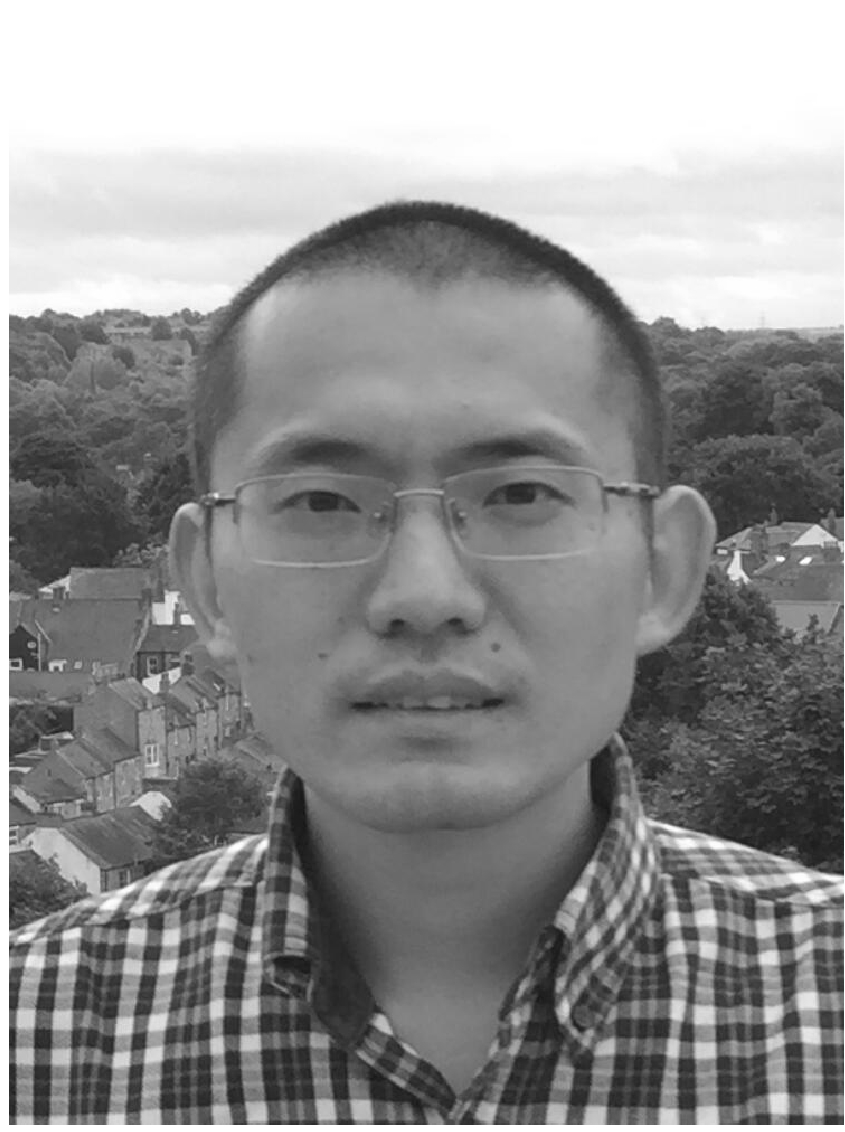}}]{Weichuan Zhang}
received the MS degree in signal and information processing from the Southwest Jiaotong University in China and the PhD degree in signal and information processing in National Lab of Radar Signal Processing, Xidian University, China. He is a research fellow at Griffith University, QLD, Australia. His research interests include computer vision, image analysis, and pattern recognition. He is a member of the IEEE.
\end{IEEEbiography}
\begin{IEEEbiography}[{\includegraphics[width=1in,height=1.25in,clip,keepaspectratio]{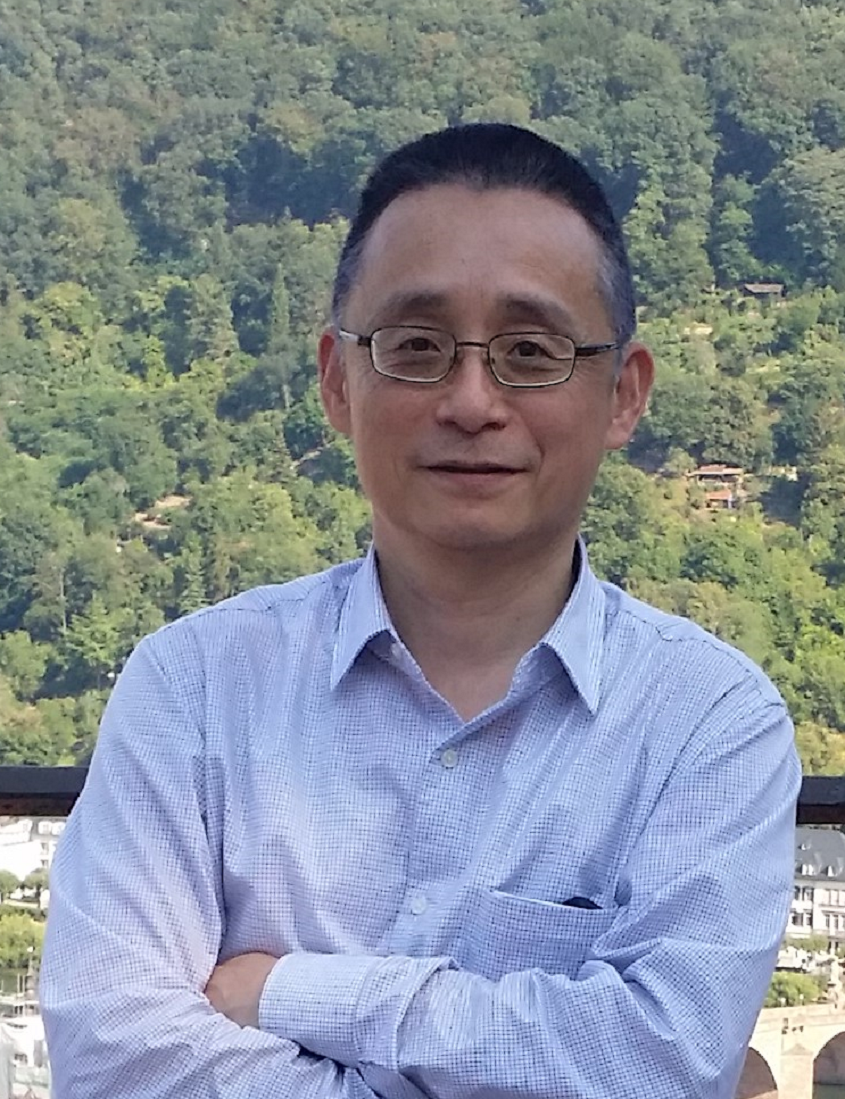}}]{Yongsheng Gao}
received the B.Sc. and M.Sc. degrees in electronic engineering from Zhejiang University, Hangzhou, China, in 1985 and 1988, respectively, and the Ph.D. degree in computer engineering from Nanyang Technological University, Singapore. He is currently a Professor with the School of Engineering and Built Environment, Griffith University, and the Director of ARC Research Hub for Driving Farming Productivity and Disease Prevention, Australia. He had been the Leader of Biosecurity Group, Queensland Research Laboratory, National ICT Australia (ARC Centre of Excellence), a consultant of Panasonic Singapore Laboratories, and an Assistant Professor in the School of Computer Engineering, Nanyang Technological University, Singapore. His research interests include smart farming, machine vision for agriculture, biosecurity, face recognition, biometrics, image retrieval, computer vision, pattern recognition, environmental informatics, and medical imaging.
\end{IEEEbiography}
\begin{IEEEbiography}[{\includegraphics[width=1in,height=1.25in,clip,keepaspectratio]{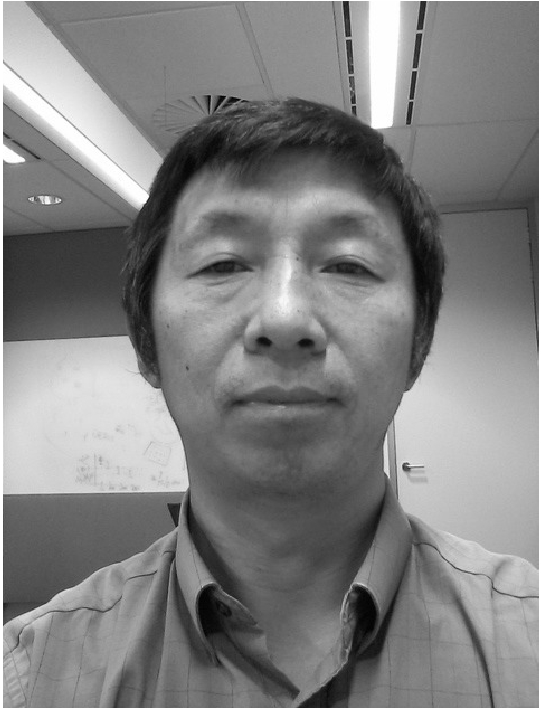}}]{Changming Sun}
received his PhD degree in computer vision from Imperial College London, London, UK in 1992. He then joined CSIRO, Sydney, Australia, where he is currently a Principal Research Scientist carrying out research and working on applied projects. He is also a Conjoint Professor at the School of Computer Science and Engineering of the University of New South Wales. He has served on the program/organizing committees of various international conferences. He is an Associate Editor of the EURASIP Journal on Image and Video Processing. His current research interests include computer vision, image analysis, and pattern recognition.
\end{IEEEbiography}

\end{document}